\documentclass[twocolumn,twoside,letter]{IEEEtran}
\usepackage{xcolor,soul,framed} %,caption
\usepackage{subfigure}
\colorlet{shadecolor}{yellow}
\usepackage[pdftex]{graphicx}
\graphicspath{{../pdf/}{../jpeg/}}
\DeclareGraphicsExtensions{.pdf,.jpeg,.png}

\usepackage[cmex10]{amsmath}
\usepackage{array}
\usepackage{mdwmath}
\usepackage{mdwtab}
\usepackage{eqparbox}
\usepackage{url}
\usepackage{amsmath}
\usepackage{amsfonts}
\usepackage{accents}
\usepackage{multirow}
\usepackage{bm}
\usepackage{algorithm}
\usepackage{algorithmic}
\IEEEoverridecommandlockouts
\begin{document}

\title{Fine-Tuning Hybrid Physics-Informed Neural Networks for Vehicle Dynamics Model Estimation~\thanks{An earlier version of this paper will be presented in part at the 2024 Modeling, Estimation, and Control Conference, Chicago, Illinois, USA, in October 27–30, 2024. ({\em Corresponding author: Kaiyan Yu}).}}

%\markboth{IEEE TRANSACTIONS ON AUTOMATION SCIENCE AND ENGINEERING,~Vol.~XX,~No.~YY, 2024}{Li~\lowercase{\textit{et al.}}: Time-Optimal Motion Planning and Control for Autonomous Vehicle}

\author{Shiming Fang\IEEEmembership{, Student Member,~IEEE,}\thanks{S. Fang and K. Yu are with the Department of Mechanical Engineering, Binghamton University, Binghamton, NY 13902 USA (email: {sfang10@binghamton.edu}; {kyu@binghamton.edu}).} %\thanks{This article has color versions of one or more figures available at		https://doi.org/10.1109/TASE.2024.XXXXXX.}, 
	and Kaiyan Yu,~\IEEEmembership{Member,~IEEE}\thanks{Source codes are available at https://github.com/Binghamton-ACSR-Lab/FTHD.}}%\thanks{Digital Object Identifier 10.1109/TASE.2024.XXXXXX}}
	\maketitle

% ====================================================================

% === ABSTRACT ====================================================================
% =================================================================================
\begin{abstract}

Accurate dynamic modeling is critical for autonomous racing vehicles, especially during high-speed and agile maneuvers where precise motion prediction is essential for safety. Traditional parameter estimation methods face limitations such as reliance on initial guesses, labor-intensive fitting procedures, and complex testing setups. On the other hand, purely data-driven machine learning methods struggle to capture inherent physical constraints and typically require large datasets for optimal performance. To address these challenges, this paper introduces the Fine-Tuning Hybrid Dynamics (FTHD) method, which integrates supervised and unsupervised Physics-Informed Neural Networks (PINNs), combining physics-based modeling with data-driven techniques. FTHD fine-tunes a pre-trained Deep Dynamics Model (DDM) using a smaller training dataset, delivering superior performance compared to state-of-the-art methods such as the Deep Pacejka Model (DPM) and outperforming the original DDM. Furthermore, an Extended Kalman Filter (EKF) is embedded within FTHD (EKF-FTHD) to effectively manage noisy real-world data, ensuring accurate denoising while preserving the vehicle's essential physical characteristics. The proposed FTHD framework is validated through scaled simulations using the BayesRace Physics-based Simulator and full-scale real-world experiments from the Indy Autonomous Challenge. Results demonstrate that the hybrid approach significantly improves parameter estimation accuracy, even with reduced data, and outperforms existing models. EKF-FTHD enhances robustness by denoising real-world data while maintaining physical insights, representing a notable advancement in vehicle dynamics modeling for high-speed autonomous racing.

\end{abstract}
 

\section{Introduction}

%Precise vehicle dynamics models are essential for developing algorithms that effectively control autonomous vehicles. These models facilitate accurate prediction and response to real-world driving conditions, particularly during high-speed and agile maneuvers, which demand precise modeling of rapid vehicle motions to enhance safety. While the kinematic models are hard to capture the dynamics aspects of the vehicle such as tire, suspension, throttling and drivetrain, this put importance of the study in dynamic models, which goes deeper into the power response and estimating the longitudinal and lateral forces on the vehicle to further predict the motion, compared to the kinematic models whose geometry parameters are easy to obtain, the sensitivity of the coefficients in the dynamic tire model (e.g. Pacejka coefficients ~(\cite{bakker1989new})) addresses more on the importance of the precise estimation, however, estimating such parameters are challenged and usually time-consuming. 

\IEEEPARstart{A}{ccurate} vehicle dynamics modeling is essential for the development of algorithms that enable effective control of autonomous vehicles, particularly in high-speed and agile driving scenarios. These models enable precise prediction and response to real-world driving conditions, which is critical for ensuring safety during rapid vehicle maneuvers. While kinematic models provide a simpler approach by focusing on geometric parameters, they fail to capture key dynamic aspects such as tire forces, suspension behavior, throttle response, and drivetrain effects. This highlights the importance of dynamic models, which delve deeper into the vehicle performance and allow for the estimation of longitudinal and lateral forces to better predict vehicle motion. A key challenge in dynamic modeling is the estimation of sensitive coefficients in tire models, such as the Pacejka coefficients~\cite{bakker1989new}, which can be both complex and time-consuming to estimate accurately.

%Data for estimating the parameters within the Magic Formula can be obtained from various sources. These include full-sized and scaled autonomous racing competitions such as the Indy Autonomous Challenge~(\cite{wischnewski2022indy}) and F1tenth~(\cite{o2019f1}), as well as simulators like BayesRace~(\cite{JainRaceOpt2020}). BayesRace uses a predefined vehicle model for pure-pursuing and MPC to generate state data.

%Traditional methods of estimating tire model encounter several limitations, including reliance on initial guesses, time-consuming fitting processes, and challenges in implementing testing setups. Due its non-linearity, Deep neural networks (DNNs) was used to capture the nonlinear effects ~(\cite{weiss2020deepracing}), this offers a simpler solution without the usage of expert testing equipment, while purely data-driven machine learning methods have gained popularity, they struggle to uncover the underlying physical constraints of the dynamics model, often relying heavily on abundant data, furthermore, the output of the purely data driven methods can also irrelevant to the actual system. 

Traditional methods for estimating tire models come with several limitations, including reliance on initial guesses, long fitting times, and challenges associated with testing setups. Deep neural networks (DNNs) have been employed to handle the non-linearity in vehicle dynamics modeling~\cite{weiss2020deepracing}, offering a simpler solution that bypasses the requirement for specialized testing equipment. Yet, purely data-driven machine learning approaches often fail to capture the underlying physical constraints of the system and are highly dependent on large datasets. Moreover, these models can produce outputs that do not reflect the real system dynamics.

%To overcome these challenges, Physics-Informed Neural Networks (PINNs) is proposed, which integrates governing equations and physical laws into neural network architectures. This integration enhances model efficiency and accuracy, making PINNs more data-efficient. However, PINNs still require rich, low-noise data to thrive, prompting researchers to explore strategies like data augmentation, transfer learning, or novel data collection techniques. Recent studies have demonstrated the effectiveness of PINNs across various scientific domains, including fluid dynamics and structural mechanics. In autonomous driving, examples are the Deep Paecjka Model (DPM) ~(\cite{kim2022physics}) and Deep Dynamics Model (DDM) ~(\cite{chrosniak2023deep}), addressing the dynamics modeling in autonomous racing with high speeds and accelerations, however, as shown in ~(\cite{chrosniak2023deep}), the DPM has several drawbacks like the reliance on sampling-based control, unconstrained parameters estimation and only simulation test done, the DDM, though shows improvements compared to the DPM which gives guard layer that constraint the estimations range, still struggles to estimate the parameters while given smaller dataset or providing noisy data for training.

To address these shortcomings, Physics-Informed Neural Networks (PINNs) have been proposed. PINNs incorporate governing equations and physical laws into neural network architectures, improving both model efficiency and accuracy, while requiring less data. Despite these advantages, PINNs still depend on high-quality, low-noise data. This has led researchers to explore techniques such as data augmentation, transfer learning, and novel data collection strategies. PINNs have been shown to be effective in various scientific domains, including fluid dynamics and structural mechanics. In the context of autonomous driving, methods like the Deep Pacejka Model (DPM)~\cite{kim2022physics} and Deep Dynamics Model (DDM)~\cite{chrosniak2023deep} have been applied to high-speed racing scenarios. However, these models have limitations. For instance, DPM relies heavily on sampling-based control and unconstrained parameter estimation, with tests only conducted in simulation environments. While DDM offers improvements by constraining estimation ranges, it still struggles with smaller datasets and noisy data during training~\cite{chrosniak2023deep}.
%Additionally, although DDM offers open-loop and closed-loop performance, utilizing Model Predictive Control (MPC) for trajectory optimization makes it challenging to assess the accuracy of the estimated magic formula parameters, as the control is optimized during the MPC process.

%Besides the requirements in large data for the supervised-PINN (driven by the input data to reveal the physic law), the unsupervised-PINN (leverage the governing equations directly) and Hybrid-PINN (combination of both methods) requires less labeled data, based on this, we present Fine-Tuning Hybrid Dynamics (FTHD) that to the best of our knowledge the first model that use fine tuning hybrid PINN model to estimate the vehicle dynamic model. Further more, we present Extended Kalman Filter FTHD (EKF-FTHD) which to the best of our knowledge the first model that use embedded extended kalman filter to estimate vehicle dynamic model. 

While supervised PINNs (which rely on input data to reveal physical laws) require large datasets, unsupervised PINNs (which leverage governing equations directly) and hybrid approaches require fewer labeled data points. Building on this, we propose the Fine-Tuning Hybrid Dynamics (FTHD) method, which, to the best of our knowledge, is the first to use a fine-tuning hybrid PINN model to estimate vehicle dynamics. Furthermore, we introduce the Extended Kalman Filter FTHD (EKF-FTHD), the first model to incorporate an embedded EKF to denoise real-world data while maintaining physical insights.

The major contributions of this paper are as follows:
\begin{enumerate}
    \item We present FTHD, a fine-tuned hybrid PINN method designed to estimate key vehicle parameters, including Pacejka tire coefficients, drivetrain coefficients, and moment of inertia. Compared to the state-of-the-art methods, FTHD achieves higher accuracy while requiring smaller datasets. 
   % - A fine-tuning PINN that use hybrid method to estimate Pajecka tire coefficients, drivetrain coefficients and moment of inertia, compare to the other recently methods, the Fine-Tuning Hybrid Dynamics (FTHD) can use smaller sized of data and give higher accuracy of the estimation.
    \item We introduce EKF-FTHD, a data processing method that embeds an EKF within the FTHD framework. This approach denoises real-world data by separating the noise from the physical signal, providing more accurate predictions. Compared to traditional smoothing methods and recent PINN models, EKF-FTHD demonstrates superior performance in revealing physical insights from noisy data, with reduced loss and improved prediction accuracy during further training. 
    %- a data processing method based on FTHD, uses embedded extended kalman filter in the model to denoise the real data, aiming to separate the noisy data into the noise part and physical part, compared to the traditional data smooth methods and PINN modeling, the EKF-FTHD is capable to reveal physical part within noisy data and shows smaller loss and higher prediction accuracy in further data training.
\end{enumerate}

Building upon our previous conference publication~\cite{FangMECC2024}, which introduced the FTHD method based on the ideal simulation data for scaled vehicles, this journal paper extends the work by proposing EKF-FTHD model to handle real-world experimental data. EKF-FTHD functions as a data-processing model that filters and denoises noisy, real-world data while retaining essential physical characteristics, allowing FTHD to achieve more accurate parameter estimation and a lower validation loss. This improvement directly addresses the limitations of previous models in managing noisy datasets. In addition, this journal paper provides further experimental results and more detailed analyses, demonstrating that the hybrid approach, even with smaller training datasets, significantly improves parameter estimation in real-world conditions. This advancement marks a major step forward in vehicle dynamics modeling for high-speed autonomous racing.

The rest of the paper is organized as follows:
In Section~\ref{sec:Related Work}, we discuss existing methods for vehicle dynamics modeling, their limitations, and recent advancements.
Section~\ref{sec:overview} outlines the problem statement and provides an overview of the vehicle model, the proposed FTHD model, and the EKF-FTHD pre-processing method for handling noisy data. The experimental results are presented in Section~\ref{sec:result}, where the performance of the FTHD and EKF-FTHD models is evaluated. Finally, the concluding remarks are summarized in Section~\ref{sec:concl}, highlighting the key contributions and future directions.

\section{Related Work}
\label{sec:Related Work}
Vehicle dynamics modeling for autonomous driving has been extensively studied, with models ranging from simple point mass representations to more complex single-track (bicycle) dynamic models (STMs) and multi-body systems. The choice of model often depends on the complexity of the driving scenario. For low-speed, non-slip conditions, kinematic models are typically used as they focus on geometric aspects. However, these models struggle to represent more complex driving conditions, especially during high-speed maneuvers where dynamic factors such as tire forces and vehicle inertia become critical. While the bicycle model strikes a balance between simplicity and accuracy in high-speed and drifting scenarios, maintaining accurate parameter estimation for this model remains a challenge~\cite{vicente2020linear}.
Several data-driven methods have been proposed to estimate vehicle model parameters. For example,~\cite{dias2014longitudinal} introduced a data-driven approach for identifying longitudinal dynamics, and~\cite{vyasarayani2011parameter} explored homotopy optimization for parameter identification in dynamic systems modeled by ordinary differential equations. However, these methods often assume simplified conditions like low speeds and time-invariant parameters, which are inadequate for the high-speed, time-dependent dynamics seen in autonomous racing, as highlighted in~\cite{chrosniak2023deep}.
%It is a widely studied topics for dynamics modeling in autonomous vehicle driving, based on the research purpose and complexity, the vehicle dynamics model can vary from a simple point mass, to single track models (bicycle model) and multi-body complex models. When the object is to study the modeling under low speed and non-slip conditions, the kinematic method is usually used, while it only consider geometry aspects and can hardly describe more complex condition, moreover, while the bicycle dynamics model hold a good balance of simplicity and accuracy in describing high speed and drifting scenarios, the accuracy of such model's parameters still hard to maintain~(\cite{vicente2020linear}). Researchers have explored several methods have been developed to learn the parameters of vehicle models. For instance, \cite{dias2014longitudinal} present a data-driven method for identifying a longitudinal dynamic model of a car, while~\cite{vyasarayani2011parameter} explore a homotopy optimization approach for parameter identification in dynamic systems modeled by ordinary differential equations. However, these methods often make assumptions such as low-speed conditions and time-unrelated parameters, which may not hold true in autonomous racing, as noted by~\cite{chrosniak2023deep}.
Alternative approaches, such as those presented by~\cite{JainRaceOpt2020} and~\cite{kabzan2019learning}, apply Gaussian Processes Regression to account for model uncertainties and enhance prediction accuracy. While these methods capture time-dependent physics, they may not always guarantee predictions that satisfy physical constraints. 

%Estimating parameters within the magic formula presents a non-linear problem, leading to the development of using deep neural networks (DNNs), which have demonstrated effectiveness in describing non-linear problems in autonomous racing (\cite{weiss2020deepracing}). The use of DNNs mitigates the complexity and high costs associated with traditional methods. However, achieving high accuracy with DNNs still necessitates large datasets, and in some cases, may produce outputs that lack utility. Moreover, purely data-driven networks like those discussed in~\cite{spielberg2019neural} and~\cite{williams2017information}  struggle to reveal the underlying physical laws of vehicle dynamics.
The estimation of parameters for tire models, such as the widely-used Pacejka Magic Formula~\cite{bakker1989new}, is inherently non-linear. Recent advancements using deep neural networks have shown promise in capturing these non-linearities, especially in autonomous racing~\cite{weiss2020deepracing}. While DNNs reduce the complexity and costs of traditional parameter estimation methods, they still require large datasets for training and often fail to capture the underlying physical laws. Furthermore, purely data-driven models, such as those described in~\cite{spielberg2019neural} and~\cite{williams2017information}, struggle to generalize to scenarios beyond the data they were trained on and can produce outputs that are not physically meaningful.
In an effort to address these challenges, \cite{hermansdorfer2021end} used a Gated Recurrent Unit (GRU)-structured neural network to potentially replace traditional physics-based STMs in simulating autonomous racing vehicles. While this approach yields more accurate predictions of vehicle dynamics within the training data, it still encounters difficulties in predicting conditions beyond the scope of the training data.

%Single-track dynamic models (STMs) in conjunction with tire models, such as the Pacejka Magic Formula~(\cite{bakker1989new}), are preferred for estimating vehicle behavior due to their simplicity and better accuracy compared to kinematics models. 

The growing interest in revealing physical laws has led to increasing research into PINNs, which combine data-driven models with the ability to uncover underlying physical laws~\cite{xu2022physics}.
For example, recent work in the field of autonomous driving, such as~\cite{koysuren2023online}, implements an online-adjusted method to estimate cornering stiffness in real-time. However, their approach assumes a linear relationship between slip angles and forces, which restricts its applicability in handling non-linear regions, especially at large slip angles. This limitation becomes critical in high-speed autonomous racing scenarios involving significant drifting. Other recent studies, such as DPM~\cite{kim2022physics} and DDM~\cite{chrosniak2023deep}, focus on estimating the magic formula using PINN-based models. However, DPM relies on sampling-based control and exhibits unconstrained Pacejka coefficient estimations, which reduces its effectiveness~\cite{chrosniak2023deep}. In contrast, DDM has demonstrated improved performance over DPM by producing results within reasonable ranges, but it suffers from a lack of robustness when trained on smaller datasets. While DDM offers both open-loop and closed-loop performance, it incorporates Model Predictive Control (MPC) for trajectory optimization. As a result, it becomes difficult to evaluate the accuracy of the estimated magic formula parameters because the control is optimized within the MPC process. Additionally, as shown in \cite{chrosniak2023deep}, when handling real-world data collected from sensors, the inherent noise from measurements introduces significant challenges. This noise leads to high losses, oscillations in the predicted states, and difficulties in model convergence, thereby obscuring the true underlying physical insights. Traditional data pre-processing methods, such as the low-pass Savitzky-Golay filter \cite{schafer2011savitzky}, can smooth the noisy data, but as curve-fitting methods, they fail to adequately reveal the physical insights, resulting in persistently high losses.

To overcome these challenges, we introduce FTHD, a novel hybrid approach that integrates PINN-based fine-tuning with enhanced noise-handling capabilities. FTHD is designed to reduce the required dataset size and improve parameter estimation by separating noise from the data, leading to more accurate predictions. To the best of our knowledge, this is the first model that employs hybrid techniques for fine-tuning vehicle dynamics modeling in high-speed autonomous racing environments. Fine-tuning, a widely used technique in machine learning, involves adapting a pre-trained model to a related task. This approach is commonly utilized across fields such as Natural Language Processing and Computer Vision, demonstrating its versatility and effectiveness.
%To address these challenges, we introduce Fine-Tuning Hybrid Dynamics (FTHD), a hybrid method of PINN-based fine-tuning model. This approach aims to reduce the dataset size and enhance parameter estimations, separating the noise data out and leading to more accurate predictions. To the best of our knowledge, it is the first model that use hybrid methods for fine-tuning vehicle dynamics modeling in high speed autonomous racing. Fine-tuning is a widely used technique in machine learning, which adapts a pre-trained model to a related task. This approach is widely utilized across various fields, including Natural Language Processing and Computer Vision. For instance, 

To further address the challenges posed by noisy data in dynamic modeling, we extended the design of the guard layer from DDM and FTHD by incorporating an EKF guard layer. This layer leverages the EKF's ability to estimate the state and predict errors in non-linear systems~\cite{ribeiro2004kalman}. The proposed EKF-enhanced FTHD, EKF-FTHD, aims to separate noisy data into two distinct components: a physical part that reflects the underlying vehicle dynamics and a noise part that can be discarded in further training. By doing so, the EKF-FTHD ensures that the physical insights of the vehicle dynamics are retained, leading to improved model accuracy.

In terms of data collection for dynamic model parameter estimation, various platforms can be utilized, including full-sized and scaled autonomous racing competitions like the Indy Autonomous Challenge~\cite{wischnewski2022indy} and F1tenth~\cite{o2019f1}. Simulators such as BayesRace~\cite{JainRaceOpt2020}, which use predefined vehicle models with MPC to generate state data, are also widely employed. These platforms provide essential datasets for training and testing the models proposed in this work.

\section{Problem Statement}
\label{sec:overview}
% Increase the row height for aesthetic spacing.
%\renewcommand{\arraystretch}{1.3}

\subsection{Vehicle Dynamics}
\label{sec:Vehicle Dynamics}
\begin{figure}[h!]
  \centering
  \includegraphics[width=0.83\linewidth]{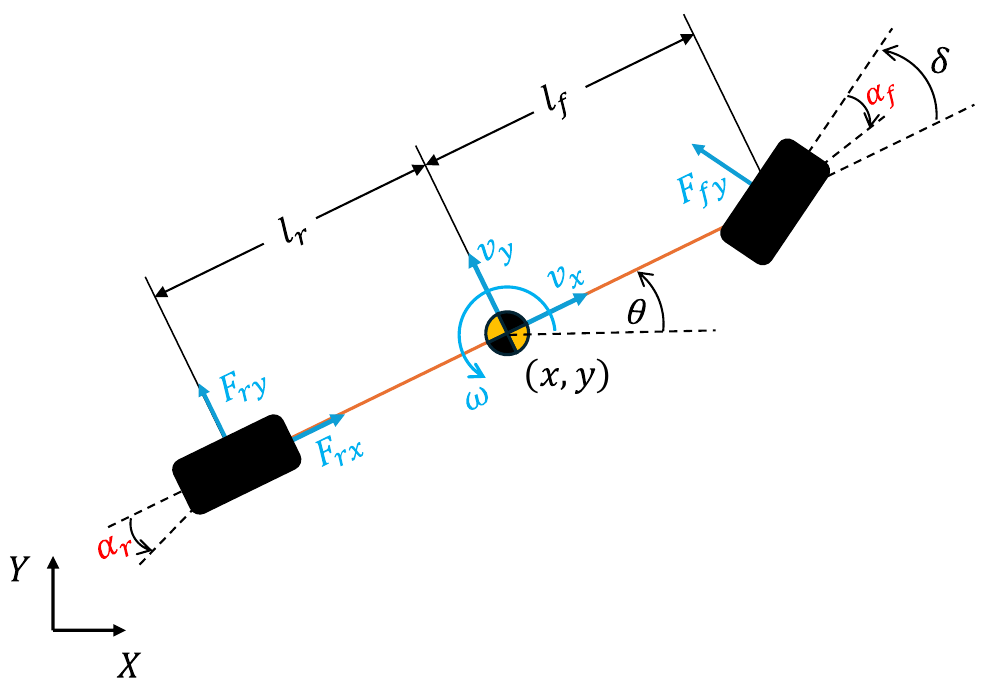}
  \vspace{-3mm}
  \caption{A schematic of the STM configuration.}
  \label{fig:SingleTrackModel}
\end{figure}

In this paper, we employ the STM, also known as the bicycle dynamic model. The STM simplifies the four-wheel structure into a two-wheel bicycle structure by assuming that the slip angles for the wheels on a particular axle are the same, effectively combining all wheels on one axle into a single wheel in the middle, as depicted in Fig.~\ref{fig:SingleTrackModel}. The state and input variables of the model are outlined in Table~\ref{tab:state_variables}.\vspace{-3mm}

\renewcommand{\arraystretch}{1.1}
\begin{table}[h!]
\caption{Notation for state and input variables}
\vspace{-3mm}
\label{tab:state_variables}
\begin{center}
\begin{tabular}{c|c}
\hline
\hline
Variables & Notation \\
\hline 
Horizontal Position (m) & \( x \) \\\hline
Vertical Position (m) & \( y \) \\\hline
Inertial Heading (rad) & \( \theta \) \\\hline
Longitudinal Velocity (m/s) & \( v_x \) \\\hline
Lateral Velocity (m/s) & \( v_y \) \\\hline
Yaw Rate (rad/s) & \( \omega \) \\\hline
Throttle (\%) & \( \mathcal{T} \) \\\hline
Steering Angle (rad) & \( \delta \) \\\hline
Change in Throttle (\%) & \( \Delta \mathcal{T} \) \\\hline
Change in Steering Angle (rad) & \( \Delta \delta \) \\
\hline
\hline
\end{tabular}
\end{center}
\end{table}
\begin{figure*}[t!]
  \centering
  \includegraphics[width=0.7\linewidth]{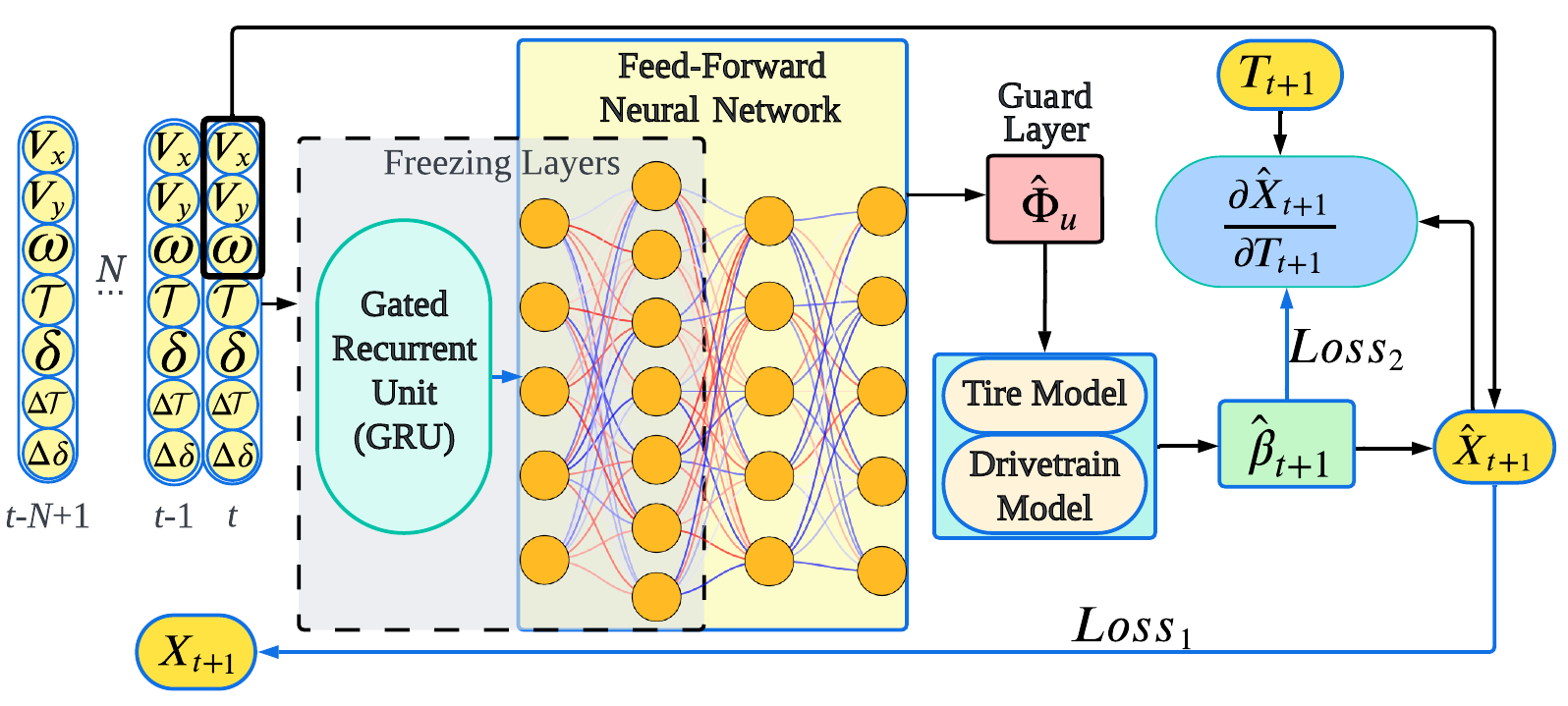}
  \caption{The structure of the proposed FTHD PINN model involves freezing layers of the pre-trained DDM and using a hybrid unsupervised loss for fine-tuning.}
  \label{fig:Fine-tuningModel}
\end{figure*}

As reported in~\cite{JainRaceOpt2020}, the longitudinal force at the rear wheel is determined by: 
\begin{equation}
\label{Drivetrain_Model}
F_{rx} = (C_{m1} - C_{m2} {v_x}^2)\mathcal{T} - C_{r0} - C_d {v_x}^2.
\end{equation}
\(C_{m1}\) and \(C_{m2}\) represent drivetrain coefficients, with \(C_{m1}\) (N) linearly related to throttle and  \(C_{m2}\) (kg/s) related to damping. \(C_{r0}\) (N) and \(C_{d}\) (kg/m) denote the rolling resistance and the drag resistance, respectively. This simplifies the drag and rolling resistance forces as \(F_d = C_d v^2_x\) and \(F_{ro} = C_{r0}\).
According to the Pacejka tire model, the slip angles $\alpha_f$ and $\alpha_r$ can be calculated as 
\begin{equation*}
\begin{aligned}
\alpha_f = \delta - \arctan\left(\frac{\omega l_f + v_y}{v_x}\right) + S_{hf},\\
\alpha_r = \arctan\left(\frac{\omega l_r - v_y}{v_x}\right) + S_{hr},
\end{aligned}
\end{equation*}
%where \(S_{hf}\) and \(S_{hr}\) represent the horizontal shift of the front and rear slip angles, respectively. Let \(S_{vf}\) and \(S_{vr}\) represent the vertical shift of the front and rear lateral forces, the lateral force at the front wheel will be $F_{yf} = S_{vf} + D_f \sin\left(C_f \arctan\left(B_f \alpha_f - E_f (B_f \alpha_f - \arctan(B_f \alpha_f))\right)\right)$ and the lateral force at the rear wheel will be $F_{yr} = S_{vr} + D_r \sin\left(C_r \arctan\left(B_r \alpha_r - E_r (B_r \alpha_r - \arctan(B_r \alpha_r))\right)\right)$, respectively. $B_i, C_i, D_i, E_i$, $i=f,r$ are the remaining Pacejka coefficients (\cite{pacejka1992magic}). \(l_f\) and \(l_r\) denote the distances from the vehicle's center of mass to the front and rear axles, respectively.
where \(S_{hf}\) and \(S_{hr}\) represent the horizontal shift of the front and rear slip angles, respectively. Let \(S_{vf}\) and \(S_{vr}\) represent the vertical shift of the front and rear lateral forces. \sloppy The lateral forces are given by $F_{fy} = S_{vf} + D_f \sin\left(C_f \arctan\left(B_f \alpha_f - E_f (B_f \alpha_f - \arctan(B_f \alpha_f))\right)\right)$ at the front wheel and $F_{ry} = S_{vr} + D_r \sin\left(C_r \arctan\left(B_r \alpha_r - E_r (B_r \alpha_r - \arctan(B_r \alpha_r))\right)\right)$~at the rear wheel.
Here, $B_i, C_i, D_i, E_i$, $i=f,r$ are the remaining Pacejka coefficients~\cite{pacejka1992magic}. \(l_f\) and \(l_r\) denote the distances from the vehicle's center of mass to the front and rear axles, respectively.
% \sloppy The lateral force at the front wheel $F_{yf} = K_f + D_f \sin\left(C_f \arctan\left(B_f \alpha_f - E_f (B_f \alpha_f - 
% \arctan(B_f \alpha_f))\right)\right)$, and the lateral force at the rear wheel
% $F_{yr} = K_r + D_r \sin\left(C_r \arctan\left(B_r \alpha_r - E_r (B_r \alpha_r - \arctan(B_r \alpha_r))\right)\right)$, where $B_i, C_i, D_i, E_i, K_i$, $i=f,r$ are the remaining Pacejka coefficients.
% \begin{table}[!htbp]
% \centering
% \caption{Vehicle Dynamics Equations and Coefficients}
% \label{tab:vehicle_dynamics}
% \begin{tabular}{@{}p{\columnwidth}@{}}
% \hline
% \textbf{Pacejka Tire Model} \\
% % \hline
% \(\alpha_f = \delta - \arctan\left(\frac{\omega l_f + v_y}{v_x} + G_f\right)\) \\
% \(\alpha_r = \arctan\left(\frac{\omega l_r - v_y}{v_x} + G_r\right)\) \\
% \(F_{yf} = \) \\ \quad \( K_f + D_f \sin\left(C_f \arctan\left(B_f \alpha_f - E_f (B_f \alpha_f - \arctan(B_f \alpha_f))\right)\right)\) \\
% \(F_{yr} = \) \\ \quad \( K_r + D_r \sin\left(C_r \arctan\left(B_r \alpha_r - E_r (B_r \alpha_r - \arctan(B_r \alpha_r))\right)\right)\) \\
% \hline
% \textbf{Drivetrain Model} \\
% % \hline
% \(F_{rx} = (C_{m1} - C_{m2} v_x^2)T - C_{r0} - C_d v_x^2\) \\
% \hline
% \end{tabular}
% \begin{tabular}{@{}p{\columnwidth}@{}}
% \textbf{Pacejka Coefficients} \\ \(B_{f/r}, C_{f/r}, D_{f/r}, E_{f/r}, G_{f/r}, K_{f/r}\) \\ 
% \textbf{Drivetrain Coefficients} \\ \(C_{m1}, C_{m2}, C_{r0}, C_d\) \\ 
% \textbf{Vehicle Geometry} \\ \(I_z, l_f, l_r, m\) \\
% \hline
% \end{tabular}
% \end{table}
At time \(t\), the system states are represented by \( \mathbf{S}_t = [x_t, y_t, \theta_t, v_{x_t}, v_{y_t}, \omega_t, \mathcal{T}_t, \delta_t] \in \mathbb{R}^8 \), and the control inputs \(U_{t}\) include changes in throttle and steering: $U_t = [\Delta \mathcal{T}_t, \Delta \delta_t]^\top \in \mathbb{R}^2$. The state equation can be obtained as follows:
\begin{equation}
\begin{aligned}
x_{t+1} &= x_t + (v_{x_t} \cos \theta_t - v_{y_t} \sin \theta_t)\Delta t,  \\
y_{t+1} &= y_t + (v_{x_t} \sin \theta_t + v_{y_t} \cos \theta_t)\Delta t, \\
\theta_{t+1} &= \theta_t + (\omega_t)\Delta t, \\
v_{x_{t+1}} &= v_{x_t} + \frac{1}{m} (F_{rx} - F_{fy} \sin \delta_t + m v_{y_t} \omega_t)\Delta t, \\
v_{y_{t+1}} &= v_{y_t} + \frac{1}{m} (F_{ry} + F_{fy} \cos \delta_t - m v_{x_t} \omega_t)\Delta t, \\
\omega_{t+1} &= \omega_t + \frac{1}{I_z} (F_{fy} l_f \cos \delta_t - F_{ry} l_r)\Delta t, \\
\mathcal{T}_{t+1} &= \mathcal{T}_t + \Delta \mathcal{T}, \quad\delta_{t+1} = \delta_t + \Delta \delta,
\end{aligned}
\label{state_equation}
\end{equation}
where \(m\) denotes the mass of the vehicle, \(I_z\) represents the moment of inertia, and \(\Delta t\) is the time step between the next time step and the current time step. 

To estimate the vehicle dynamics, similar to DDM, we categorize all coefficients of the model into two groups: \(\Phi_k = [l_r, l_f, m]\), which are readily available or easy to measure, and \(\Phi_u = [B_i, C_i, D_i, E_i, S_{hi}, S_{vi}, C_{m1}, C_{m2}, C_{r0}, C_d, I_z]\), where $i=f,r$, which are more challenging to measure directly and will be estimated by the model. Let \(\Phi = \Phi_k \cup \Phi_u\). This paper employs a hybrid method to fine-tune the DDM, assuming that each estimated coefficient falls within a predefined nominal range \(\underaccent{\rule{.4em}{.8pt}}{\Phi}_u \leq \Phi_u \leq \accentset{\rule{.4em}{.8pt}}{\Phi}_u\), where $\underaccent{\rule{.4em}{.8pt}}{\Phi}_u$ and $\accentset{\rule{.4em}{.8pt}}{\Phi}_u$ represent the lower and upper bounds, respectively.
Additionally, the modeling approach assumes certain conditions: the vehicle operates on a 2D surface, disregarding the influence of the z-axis; any delay in the action commands is neglected, i.e., the inputs are applied to the model without any delays.

\subsection{FTHD PINN Model}

In the proposed FTHD model, similar to the models defined by DDM~\cite{chrosniak2023deep} and DPM~\cite{kim2022physics}, with the given \(X_t = [v_{x_t}, v_{y_t}, \omega_t, \mathcal{T}_t, \delta_t] \in \mathbb{R}^5\) and \(U_t = [\Delta \mathcal{T}, \Delta \delta] \in \mathbb{R}^2\), the evolution of the position states \(x_t, y_t, \theta_t\) can be calculated using Eq.~(\ref{state_equation}).

During the pre-training phase, like DDM, FTHD collects \(N\) continuous time steps (\(N \geq 1\)) as the input \(\mathbf{X}_\text{input} = [X_{t-N+1},...,X_{t}]\) from the training dataset. Subsequently, it predicts the estimated state \(\hat{X}_{t+1} = [\hat{v}_{x_{t+1}},\hat{v}_{y_{t+1}},\hat{\omega}_{t+1}]\) at time \(t+1\), which is then compared with the label \(X_{t+1} = [v_{x_{t+1}},v_{y_{t+1}},\omega_{t+1}]\) using Mean-Square-Error (MSE) loss for backpropagation:
\begin{equation*}
    Loss_1 = \text{MSE}(\hat{X}_{t+1},X_{t+1}).
\end{equation*}

After the pre-training phase, the entire state dictionary, denoted as \(\Theta = \{W,b\}\), is selected for the hybrid fine-tuning process. Here, 
\(W\) and \(b\) represent the weights and biases of the model, respectively. %During the fine-tuning process, we freeze \(n\) selected layers of the model, where $n$ varies based on the total number of hidden layers and make sure at least one layer (exclude activation layer) is still activated. 
During the fine-tuning process, we selectively freeze $n$ layers of the model, where $n$ is determined based on the total number of hidden layers in the architecture. Importantly, we ensure that at least one hidden layer (excluding activation layers) remains active and unfrozen to allow for continued learning and adaptation. The number of frozen layers $n$ is chosen to balance stability and adaptability during the fine-tuning phase, ensuring the model can still adjust to new data while preserving key features learned during pre-training. The gradients of the frozen parameters, denoted as \(\Theta_f=\{W_f, b_f\}\), are no longer updated, while only the active parameters, 
\(\Theta_a = \{W_a, b_a\}\), participate in backpropagation. This strategy allows the model to retain the knowledge captured in the frozen layers while adapting the active layers to the new data, improving performance with a smaller training dataset. %This means that the gradients of the frozen states, denoted as \(\Theta_f=\{W_f, b_f\}\), will no longer be updated, and only the active states, denoted as \(\Theta_a = \{W_a, b_a\}\), will participate in backward propagation.
This process is illustrated in Fig.~\ref{fig:Fine-tuningModel}.

Compared to the DDM, the training dataset for the proposed model includes an additional time input, \(T_{t+1}\) alongside \(X_t\) and \(U_t\), resulting in a structured dataset \(\mathbf{D} = [[X_1, U_1, T_2], \dots, [X_N, U_N, T_{N+1}]]\). This augmentation is inspired by the time constraint loss in physically constrained neural networks introduced by \cite{liu2021dual}, where a time partial differential equation (PDE) serves as one of the unsupervised loss functions. In the domain of autonomous driving, this approach is relevant due to the relationship between velocity derivatives and acceleration, as inferred from the model's output:
\begin{equation*}
    \hat{X}_{t+1}=[\hat{v}_{x_{t+1}}, \hat{v}_{y_{t+1}}, \hat{\omega}_{t+1}]=f(X_t, U_t, \Phi_k, \Phi_u, T_{t+1}).
\end{equation*}
Using input time \(T_{t+1}\) and the intermediate estimated acceleration \(\hat{\beta}_{t+1} = [a_{x_{t+1}}, a_{y_{t+1}}, \dot{\omega}_{t+1}]\), the unsupervised loss is defined as:
\begin{equation*}
    Loss_2 = \text{MSE}\left(\frac{\partial \hat{X}_{t+1}}{\partial T_{t+1}}, \hat{\beta}_{t+1}\right).
\end{equation*}
$Loss_2$ serves as an unsupervised loss that is unrelated to the label $X_{t+1}$. This loss function aims to ensure that, given the input $X_t$ and assuming the time interval $\Delta t$ is small enough, the model outputs an estimated  $\hat{X}_{t+1}$ such that the difference  \(\hat{X}_{t+1}-\hat{X}_t\) is close to the estimated acceleration $\hat{\beta}$. Without providing the label, the estimated $\hat{X}_{t+1}$ could vary freely, but it still satisfies the dynamic motion constraints.

%Following the approach of~\cite{lew2022physics} and \cite{liu2021dual}, when dealing with the hybrid loss of the model, two weights \(w_1\) and \(w_2\) are introduced such that \(w_1 + w_2 = 1\). 
%In order to have the fine-tuned model working with specific vehicle, $Loss_1$ needs to be considered more than $Loss_2$, as $Loss_2$ is used to make the model get closer to a generalized vehicle model where the derivative of local longitudinal and lateral speeds need to be the same with the longitudinal and lateral accelerations. By adding $Loss_2$ to it, the model's parameters after the DDM gets additional curvature for updating and are able to converge more into the same local minima even the given dataset size is reduced. This ensures that the model optimizes for both specific vehicle characteristics and generalized physical laws.
Following the approach of~\cite{lew2022physics} and \cite{liu2021dual}, when handling the hybrid loss of the model, two weights, \(w_1\) and \(w_2\), are introduced, where \(w_1 + w_2 = 1\).  In this framework, $Loss_1$ is prioritized to ensure the fine-tuned model effectively aligns with the dynamics of a specific vehicle, as it is directly related to vehicle-specific characteristics. On the other hand, $Loss_2$ serves to generalize the model, ensuring that the derivative of local longitudinal and lateral speeds corresponds to the longitudinal and lateral accelerations, thereby refining the underlying physics.
By incorporating $Loss_2$, the model parameters receive additional curvature updates after the DDM, allowing for improved convergence towards a shared local minimum, even when the dataset size is reduced. This approach ensures that the model retains both specific vehicle dynamics and generalizable physical insights, optimizing performance with limited data.
The total loss of the model, $Loss_\text{total}$, is then calculated as a weighted sum of $Loss_1$ and $Loss_2$, expressed as:
\begin{equation}
	\label{Total_Loss}
	Loss_\text{total} = w_1 \cdot Loss_1 + w_2 \cdot Loss_2.
\end{equation}
Eq.~(\ref{Total_Loss}) combines the supervised loss \(Loss_1\) (driven by the training labels and the model outputs) with the unsupervised loss \(Loss_2\) (driven by the PDE constraints of the model's internal features, without relying on labels). This approach assumes that when the loss driven by labeled data is sufficiently low but struggles to further decrease the validation error, integrating the unsupervised loss provides additional constraints to the model, allowing for more opportunities for the model's state to update. By using this hybrid method with adaptive weights, which combines supervised and unsupervised learning, it becomes possible to train the model with less data while still achieving higher accuracy compared to using solely supervised PINNs.

We incorporate the physics guard layer, designed for DDM, which proves beneficial as it restricts the output Pacejka coefficients within a predefined nominal boundary \([\underaccent{\rule{.4em}{.8pt}}{\Phi}_u , \accentset{\rule{.4em}{.8pt}}{\Phi}_u]\), ensuring that the resulting relationships between lateral force and slip angle are always physically consistent. %ensuring that the resulting \(F_{yf/yr}\)-\(\alpha_{f/r}\) relationships are always rational. 
This is achieved through:
$\hat{\Phi}_u = \sigma (z) \cdot (\accentset{\rule{.4em}{.8pt}}{\Phi}_u - \underaccent{\rule{.4em}{.8pt}}{\Phi}_u) + \underaccent{\rule{.4em}{.8pt}}{\Phi}_u$, %where $z$ represents the elements inside last hidden layer, restricted by the size of $\Phi_u$, %which is then passed through sigmoid activation layers and scaled by the preset range into the estimated $\hat{\Phi}_u$.
where $z$ represents the elements in the last hidden layer, constrained by the size of $\Phi_u$. These are then passed through sigmoid activation layers and scaled by a preset range to yield the estimated $\hat{\Phi}_u$.

However, unlike the pure-data driven DNN, the prediction \(\hat{X}_{t+1}\) is derived from the input state and the estimated Pacejka coefficients based on the STM vehicle dynamics and Eq.~(\ref{state_equation}), the convergence of \(Loss_1\) may be influenced by the backpropagation of the guard layer. Even though the guard layer ensures that the estimated Pacejka coefficients fall within the range, it can still be challenging for the model to find the optimal states that satisfy all drifting conditions. This challenge can be attributed to the local minima problem, as discussed in~\cite{swirszcz2016local}, which demonstrates that a sigmoid-based neural network can lead to suboptimal local minima. Since the guard layer of DDM employs a sigmoid function and outputs each of the Pacejka coefficients independently, if one of them reaches a local minimum, the sigmoid function may struggle to further update it. These local minimum coefficients can then influence other coefficients, especially when they are combined in the tire model and the drivetrain model in Eq.~(\ref{Drivetrain_Model}), particularly as the slip angle increases and the force relationship becomes nonlinear~\cite{bakker1989new}. Moreover, when the size of the training dataset is reduced, if one or more Pacejka coefficients reach local minima, the entire model may converge to a local minimum, making it challenging to further reduce the loss if no new data is added to the training process.

Therefore, the proposed FTHD utilizes a fine-tuning model and unsupervised method to address this issue. If the model is reaching or approaching a local minimum, freezing some layers and adding new loss functions to the loss space could help update coefficients whose gradients do not change. The use of a hybrid method could further reduce the training dataset while maintaining accuracy compared to supervised-only PINNs.

\subsection{EKF-FTHD Pre-Processing Method}
As discussed in DDM~\cite{chrosniak2023deep}, when processing data collected from real-world scenarios like the Indy Autonomous Challenge, noise poses a significant challenge for achieving convergence comparable to simulation-based models. This is largely due to sensor-based data, which relies on physical measurements rather than computed values. As a result, the real data often exhibits oscillations, even when the car operates at nearly stable speeds. These oscillations introduce noise, degrading the model's performance. While traditional smoothing methods, such as the Savitzky-Golay filter~\cite{schafer2011savitzky}, can improve convergence by smoothing the data, they overlook the physical relationships between variables such as \(X_t\) and \(U_t\), potentially leading to feature loss and residual errors.

%As shown in DDM~\cite{chrosniak2023deep}, while dealing with the data collected by the real Indy Autonomous Challenge, due to the noise problem, it is hard to have the DDM converge as accurate as in simulation, this is due to the reason that the data from the sensor is based on measurements instead of calculation, therefore the real data is kept oscillating even the real car is hardly at perfect stable speed, this highly influence the performance of the model, though using traditional smooth method like applying Sacitzky-Golay filter~\cite{schafer2011savitzky} can indeed smooth the data and improve the convergence, however, the smooth procedure does not consider the physics relationship between \(X_t\) and \(U_t\), causing the lost of features and residual errors after smoothing.

\begin{figure*}[t!]
	\centering
	\includegraphics[width=0.8\linewidth]{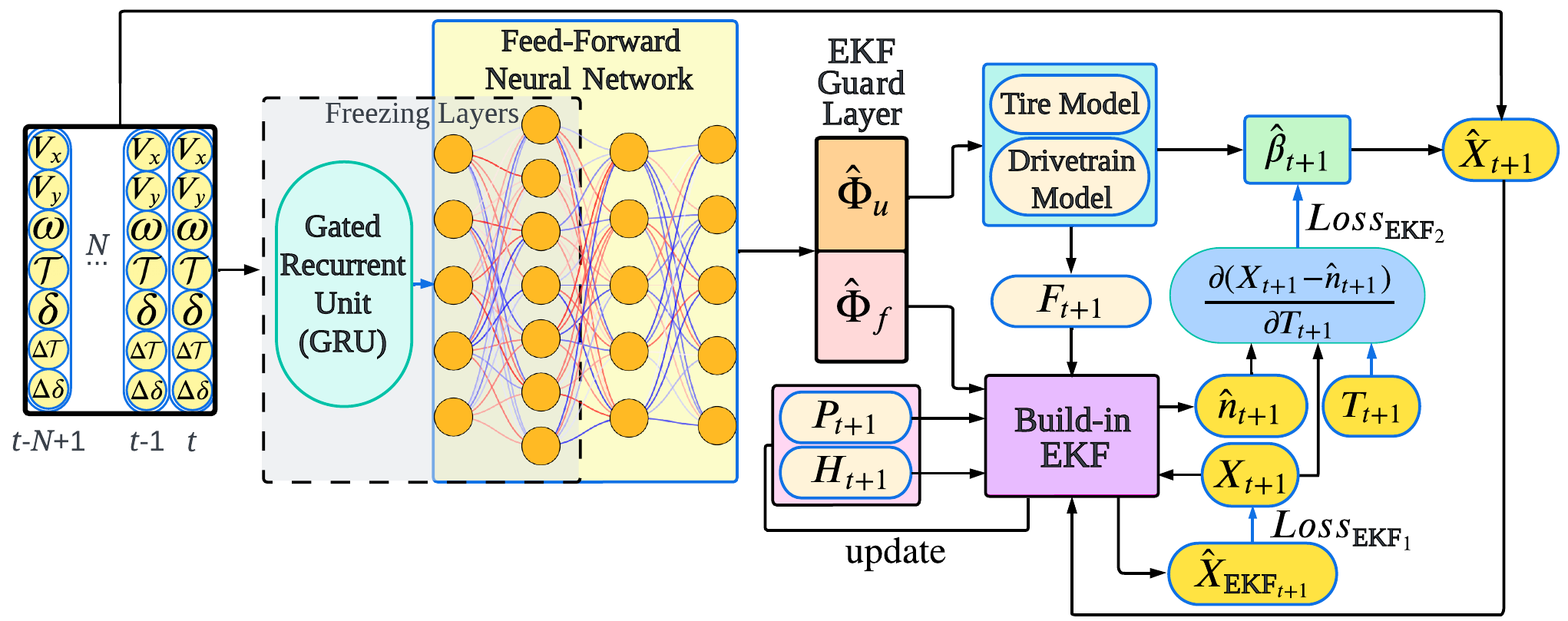}
	\caption{The architecture of the proposed EKF-FTHD PINN model for data processing. This model incorporates an embedded EKF with dynamically adjustable covariance matrices to filter out noise from sensor measurements. }
	\label{fig:EKF_Fine-tuningModel}
\end{figure*}

To address this, we propose the EKF-FTHD pre-processing method, which leverages EKF's ability to estimate states in nonlinear systems. Table~\ref{tab:ekf_state_variables} summarizes the variables used in the proposed method. Here, \(\hat{X}_{t+1}\) represents the estimated state output by the model, while \(X_{t+1}\) denotes the measurements obtained from the sensor at the time step $t+1$. 
\renewcommand{\arraystretch}{1.3}
\begin{table}[h!]
	\caption{Notation for variables used in the EKF-FTHD model}
	\label{tab:ekf_state_variables}
	\vspace{-3mm}
	\begin{center}
		\begin{tabular}{c|c}
			\hline
			\hline
			Variables & Notation \\
			\hline 
			Model state vector at time $t+1$ & \( \hat{X}_{t+1} \) \\\hline
			Filtered state vector at time $t+1$ & \( \hat{X}_{\text{EKF}_{t+1}} \) \\\hline
			Measurement at time $t+1$ & \( X_{t+1} \) \\\hline
			Nonlinear state transition function & \( f(\cdot) \) \\\hline
			Nonlinear measurement function & \( h(\cdot) \) \\\hline
			Jacobian of \(f\) w.r.t. \(X_{t+1}\) & \( \bm{F}_{t+1} \) \\\hline
			Jacobian of \(h\) w.r.t. \(\hat{X}_{t+1}\) & \( \bm{H}_{t+1} \) \\\hline
			Covariance matrix at time $t+1$ & \( \bm{P}_{t+1} \) \\\hline
			Process noise covariance & \( \bm{Q}_{t+1} \) \\\hline
			Measurement noise covariance & \( \bm{R}_{t+1} \) \\
			\hline
			\hline
		\end{tabular}
	\end{center}
\end{table}

Based on the FTHD model shown in Fig.~\ref{fig:Fine-tuningModel}, we incorporate an EKF guard layer embedded with a covariance matrix to denoise the raw state \(X_{t+1} = [v_{x_{t+1}},v_{y_{t+1}},\omega_{t+1}]\), splitting it into a physical component and a noise component: \(X_{t+1} = X_{\text{EKF}_{t+1}} \cup \epsilon\). The estimated parameters include \(\Phi_{\text{EKF}} = \Phi_u \cup \Phi_{f}\), where \(\Phi_f = [q_j,r_j]\) and \(j = v_x, v_y, \omega\). Each coefficient is assumed to lie within a nominal range: \(\underaccent{\rule{.4em}{.8pt}}{\Phi}_f \leq \Phi_f \leq \accentset{\rule{.4em}{.8pt}}{\Phi}_f\), with $\underaccent{\rule{.4em}{.8pt}}{\Phi}_f$ and $\accentset{\rule{.4em}{.8pt}}{\Phi}_f$ represent the lower and upper bounds, respectively. $q_j$ and $r_j$ are elements of the noise covariance matrices $\bm{Q}_t$ and $\bm{R}_t$, respectively. $\bm{Q}_t$ represents the covariance of the process noise during the state prediction step of the EKF, and $\bm{R}_t$ represents the covariance of the measurement noise, which accounts for uncertainty during the update step of the EKF. For simplification, both $\bm{Q}_t$ and $\bm{R}_t$ are assumed to be diagonal matrices. This assumption reduces the computational complexity, as it implies that noise variables are independent, and their covariances are zero, resulting in:
%In the proposed EKF-FTHD (Extended Kalman Filter Fine-tuning Hybrid Dynamics) pre-processing method, inspired by the ability of EKF in handling nonlinear systems to estimate state at each time steps, based on the FTHD model shown in Fig.~\ref{fig:Fine-tuningModel}, we modified the guard layer with an embedded EKF covariance matrix i.e. EKF guard layer, aiming to denoise the original noisy state \(X_{t+1} = [v_{x_{t+1}},v_{y_{t+1}},\omega_{t+1}]\) into a physical part and noise part, i.e. \(X_{t+1} = X_{\text{EKF}_{t+1}} \cup \epsilon\). The new estimated parameters will be \(\Phi_{\text{EKF}} = \Phi_u \cup \Phi_{f}\), \(\Phi_f = [Q_j,R_j]\) where \(j = v_x, v_y, \omega\), assuming that each estimated coefficient falls within a predefined nominal range \(\underaccent{\rule{.4em}{.8pt}}{\Phi}_f \leq \Phi_f \leq \accentset{\rule{.4em}{.8pt}}{\Phi}_f\), where $\underaccent{\rule{.4em}{.8pt}}{\Phi}_f$ and $\accentset{\rule{.4em}{.8pt}}{\Phi}_f$ represent the lower and upper bounds, respectively. $Q_j$ and $R_j$ are the elements in the $\bm{Q}_t$ and $\bm{R}_t$ matrix, $\bm{Q}_t$ represents the covariance of the process noise during the state prediction step of the EKF, $\bm{R}_t$ represents the covariance of the measurement noise which accounts for the uncertainty during the update step of the EKF. For simplification, $\bm{Q}_t$ and $\bm{R}_t$ use the diagnal matrix form: 
\begin{equation*}
\begin{aligned}
\bm{Q}_t &=
\begin{bmatrix}
    q_{v_x}&0&0\\
    0&q_{v_y}&0\\
    0&0&q_{\omega}
\end{bmatrix}_t, & \bm{R}_t &=
\begin{bmatrix}
    r_{v_x}&0&0\\
    0&r_{v_y}&0\\
    0&0&r_{\omega}
\end{bmatrix}_t.
\end{aligned}
\label{eq:Q_matrix}
\end{equation*}
Here, $q_{v_x}$, $q_{v_y}$, $q_{\omega}$ represent the process noise covariances associated with the longitudinal velocity, lateral velocity, and yaw rate, respectively. Similarly, 
$r_{v_x}$, $r_{v_y}$, $r_{\omega}$ represent the measurement noise covariances for these state variables. 
The filter's parameters are bounded based on how much the noise in the original data affects the physical dynamics. 

The measurement residual is defined as:
\begin{equation}
\label{eq:measurement_residual}
    \hat{Y}_{t+1} = X_{t+1} - \hat{X}_{t+1}.
\end{equation}
The state function \(f(\cdot)\), used to compute \(v_x\), \(v_y\), and \(\omega\), follows the equations in~(\ref{state_equation}). The Jacobian matrix of \(f(\cdot)\) is expressed as:
\begin{equation*}
\begin{aligned}
\bm{F}_{t+1} &=
\begin{bmatrix}
    1&\Delta t\omega_{t+1}&0\\
    -\Delta t\omega_{t+1}&1&0\\
    0&0&1
\end{bmatrix}.
\end{aligned}
\label{jacobian_state_equation}
\end{equation*}
Since the sensor data provides direct measurements, both the measurement function \(h(\cdot)\) and the measurement matrix \(\bm{H}_{t+1}\) are identity matrices in this case. The structure of the model is shown in Fig.~\ref{fig:EKF_Fine-tuningModel}, where the output matrices \(\bm{Q}_{t+1}\) and \(\bm{R}_{t+1}\) from the network along with the calculated \(\bm{F}_{t+1}\) are fed into the model’s built-in EKF. Given an initial covariance matrix \(\bm{P}_{t}\), representing the estimation error in the state vector, the predicted covariance matrix \(\bm{P}_{t+1|t}\) is updated using 
\(\bm{F}_{t+1}\) and \(\bm{Q}_{t+1}\):
\begin{equation}
\label{eq:predict_P_update}
    \bm{P}_{t+1|t} = \bm{F}_{t+1} \times \bm{P}_{t} \times \bm{F}_{t+1}^\top + \bm{Q}_{t+1}.
\end{equation}
The innovation covariance is defined as: 
\begin{equation*}
\label{eq:innovation_covariance}
\bm{S}_{t+1} = \bm{H}_{t+1} \times \bm{P}_{t+1|t} \times \bm{H}_{t+1}^\top + \bm{R}_{t+1}.
\end{equation*}
Thus, the kalman gain is calculated as:
\begin{equation}
\label{eq:kalman_gain}
    \bm{K}_{t+1} = \bm{P}_{t+1|t} \times \bm{H}_{t+1}^\top \times \bm{S}_{t+1}^{-1}.
\end{equation}
Using Eq.~(\ref{eq:measurement_residual}) and Eq.~(\ref{eq:kalman_gain}), the noise estimate \(\hat{n}_{t+1}\) becomes:
\begin{equation*}
    \label{eq:ekf_noise}
    \hat{n}_{t+1} = \bm{K}_{t+1} \times \hat{Y}_{t+1},
\end{equation*}
and the state estimate \(\hat{X}_{\text{EKF}_{t+1}}\) is updated as follows based on Eq.~(\ref{eq:predict_P_update}) and Eq.~(\ref{eq:kalman_gain}):
\begin{equation*}
    \label{eq:ekf_state_update}
    \hat{X}_{\text{EKF}_{t+1}} = \hat{X}_{t+1} + \bm{K}_{t+1} \times \hat{Y}_{t+1}.
\end{equation*}
According to Eq.~(\ref{eq:predict_P_update}) and Eq.~(\ref{eq:kalman_gain}), the covariance matrix \(\bm{P}_{t+1}\) is updated by
\begin{equation*}
    \label{eq:P_update}
    \bm{P}_{t+1} = (\bm{I} - \bm{K}_{t+1} \times \bm{H}_{t+1}) \times \bm{P}_{t+1|t},
\end{equation*}
where \(\bm{I} \in \mathbb{R}^{3 \times 3}\) denotes the $3\times3$ identity matrix. Once the data is filtered, the supervised loss is defined as:
\begin{equation*}
    % \label{eq:ekf_supervised_loss}
    Loss_{\text{\text{EKF}}_1} = \text{MSE}(\hat{X}_{\text{EKF}_{t+1}}, X_{t+1}).
\end{equation*}

Similar to FTHD, with the given time vector \(T_{t+1}\), along with the intermediate acceleration vector \(\hat{\alpha}_{t+1} = [a_{x_{t+1}}, a_{y_{t+1}}, \dot{\omega}_{t+1}]\), the unsupervised loss is defined as: 
\begin{equation*}
\label{eq:ekf_unsuper_loss}
    Loss_{\text{EKF}_2} = \text{MSE}\left(\frac{\partial (X_{t+1} - \hat{n}_{t+1})}{\partial T_{t+1}}, \hat{\alpha}_{t+1}\right).
\end{equation*}
This unsupervised loss formulation assumes that the noisy data disrupts the convergence of the governing differential equations. That is, when noisy data is provided, the governing differential equations from velocity to acceleration are challenging to converge due to the oscillatory behavior of the data. However, after separating the noise from the data, the remaining physical component should better satisfy the governing equations.
%Eq.~(\ref{eq:ekf_unsuper_loss}) is designed based on the following assumption: When noisy data is provided, the governing differential equations from velocity to acceleration are challenging to converge due to the oscillatory behavior of the data. However, once the noise in the original data is estimated and separated by the model, the remaining part that satisfies the governing differential equations will lead to a more accurate model. 
%Given the weights \(w_1\) and \(w_2\), where \(w_1 + w_2 = 1\), the hybrid total loss of the EKF-FTHD model is defined as:
%while given noisy data, the governing differential equations from velocity to acceleration is hard to converge due to the oscillated evolution of the data, once the original data's estimated noise is separated by the model, the closer the remained part is satisfied the governing differential equations, the more accurate the model will be, given the weights \(w_1\) and \(w_2\) where \(w_1 + w_2 = 1\), the hybrid total loss of the EKF-FTHD will be:
Therefore, the hybrid total loss for the EKF-FTHD is calculated with the weights \(w_1\) and \(w_2\) such that \(w_1 + w_2 = 1\):
\begin{equation*}
    Loss_{\text{EKF}_\text{total}} = w_1 \cdot Loss_{\text{EKF}_1} + w_2 \cdot Loss_{\text{EKF}_2}.
\end{equation*}

\begin{figure}[t!]
  \centering
  \includegraphics[width=0.9\linewidth]{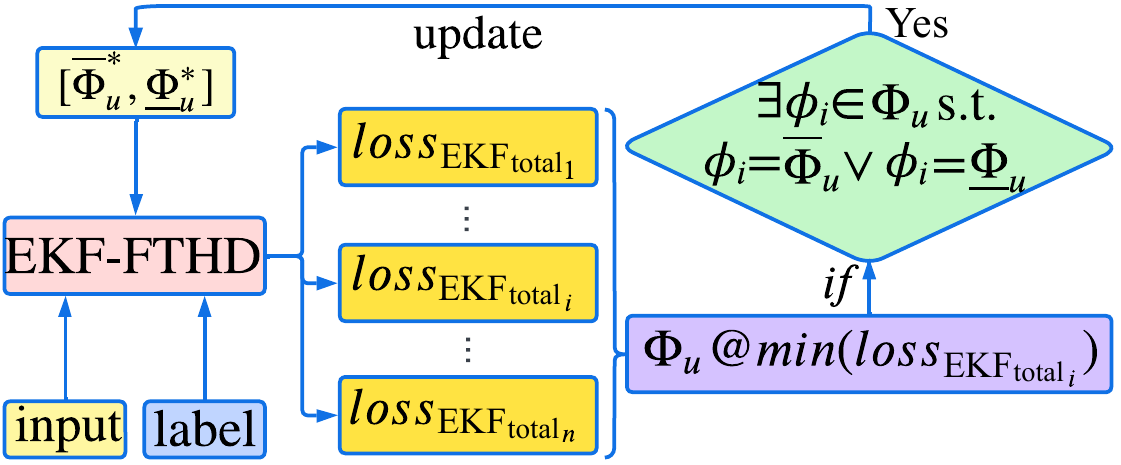}
  \caption{Illustration of the process for updating the preset coefficient ranges using the EKF-FTHD method.}
  \label{fig:EKF_parameter_adjustment}
\end{figure}

By integrating the EKF within the FTHD model and adjusting the noise covariance matrices, the EKF-FTHD effectively separates the data into physical and noise components, which contrasts with the DDM approach. The separated physical part \(X_{\text{EKF}_{t+1}}\) decreases the impact of noise and focuses more on the vehicle's dynamics. Furthermore, the range of the estimated parameters significantly influences the training process. Specifically, smaller ranges tend to yield better convergence. However, without access to the ground truth of real data, determining the optimal range is challenging. Due to the oscillatory nature of the parameters, even if the preset ranges do not align with the true values, the DDM may still estimate an intermediate value within the specified range. This introduces ambiguity in assessing whether the result from DDM is indeed the best estimate. Our experiments indicate that with the original DDM preset range \([\underaccent{\rule{.4em}{.8pt}}{\Phi}_u , \accentset{\rule{.4em}{.8pt}}{\Phi}_u]\) the model outputs coefficients that hit the boundary of the range. By adjusting the range to 
\([\underaccent{\rule{.4em}{.8pt}}{\Phi}_u^* , \accentset{\rule{.4em}{.8pt}}{\Phi}_u^*]\), both DDM and EKF-FTHD achieve smaller loss values and better convergence. Fig.~\ref{fig:EKF_parameter_adjustment} shows the process of updating the preset coefficient ranges, where \(\phi_i\) represents each parameter of \(\Phi_u\) defined in Section~\ref{sec:Vehicle Dynamics}.

%Besides the improvements of FTHD, by incorporating Extended Kalman Filter inside the model whose noise covariance matrix are adjusted accordingly, the EKF-FTHD ensures the model to separate the whole dataset into two part which compared to the DDM method, the separated physical part \(X_{\text{EKF}_{t+1}}\) decreases the influence from the noise, trying to reflect more on the physical side of the vehicle dynamic model, furthermore, the range of the estimated parameters influence a lot on the training i.e. the smaller range gives better convergence, without given the ground truth of real data, it is hard to determine which range should be decided, due to the parameters' oscillation problem, even the preset ranges do not lay on the true value, the DDM will still guess an intermediate value within the range, which makes it hard to judge whether it is the best value, our experiments shows that after the training, while using the preset range \([\underaccent{\rule{.4em}{.8pt}}{\Phi}_u , \accentset{\rule{.4em}{.8pt}}{\Phi}_u]\) from DDM, the best fitting model will output the coefficients values that hit the boundary of the range, by adjusting the range to \([\underaccent{\rule{.4em}{.8pt}}{\Phi}_u^* , \accentset{\rule{.4em}{.8pt}}{\Phi}_u^*]\) that keeps the coefficients evolving, both the DDM and EKF-FTHD could get a smaller loss and better convergence. Fig.~(\ref{fig:EKF_parameter_adjustment}) shows the process of updating the preset coefficients ranges, where \(\phi_i\) represents each of the parameters of \(\Phi_u\) defined in Sec.~\ref{sec:Vehicle Dynamics}.

After completing EKF-FTHD training, using hyperparameter tuning similar to that in the FTHD process, we select the model with the lowest loss. This pre-process model, along with the corresponding 
\(X_{\text{EKF}_t}\), is used as the new dataset, with adjusted coefficient ranges \([\underaccent{\rule{.4em}{.8pt}}{\Phi}_u^* , \accentset{\rule{.4em}{.8pt}}{\Phi}_u^*]\). To validate the effectiveness of this dataset and the adjusted ranges, we train the FTHD model following a similar simulation process with \(X_{\text{EKF}_t}\) and \([\underaccent{\rule{.4em}{.8pt}}{\Phi}_u^* , \accentset{\rule{.4em}{.8pt}}{\Phi}_u^*]\) as the dataset and coefficients ranges, demonstrating the EKF-FTHD method's advantages in data processing and coefficient adjustment. We also compare the DDM model trained with \(X_{\text{EKF}_t}\) and the adjusted range \([\underaccent{\rule{.4em}{.8pt}}{\Phi}_u^* , \accentset{\rule{.4em}{.8pt}}{\Phi}_u^*]\) to the original DDM from~\cite{chrosniak2023deep}. Results and discussions are presented in the next section.

%After the training process of EKF-FTHD, with the use of hyperparameter training similar to the simulation process, we choose the model with the lowest loss as the pre-process model and the corresponded \(X_{\text{EKF}_t}\) as the new dataset with the adjusted range \([\underaccent{\rule{.4em}{.8pt}}{\Phi}_u^* , \accentset{\rule{.4em}{.8pt}}{\Phi}_u^*]\), in order to prove the accuracy of the new dataset and preset range, we train the prediction model FTHD follow the similar process of the simulation with \(X_{\text{EKF}_t}\) and \([\underaccent{\rule{.4em}{.8pt}}{\Phi}_u^* , \accentset{\rule{.4em}{.8pt}}{\Phi}_u^*]\) as the dataset and coefficients range, to further prove the advance of the EKF-FTHD method in the data processing and coefficients range adjustment, we train the DDM model with the \(X_{\text{EKF}_t}\) and \([\underaccent{\rule{.4em}{.8pt}}{\Phi}_u^* , \accentset{\rule{.4em}{.8pt}}{\Phi}_u^*]\) to compare with the original DDM~\cite{chrosniak2023deep}, results and discussions are shown in next section.

\begin{figure}[t!]
	\centering
	\includegraphics[width=0.75\linewidth]{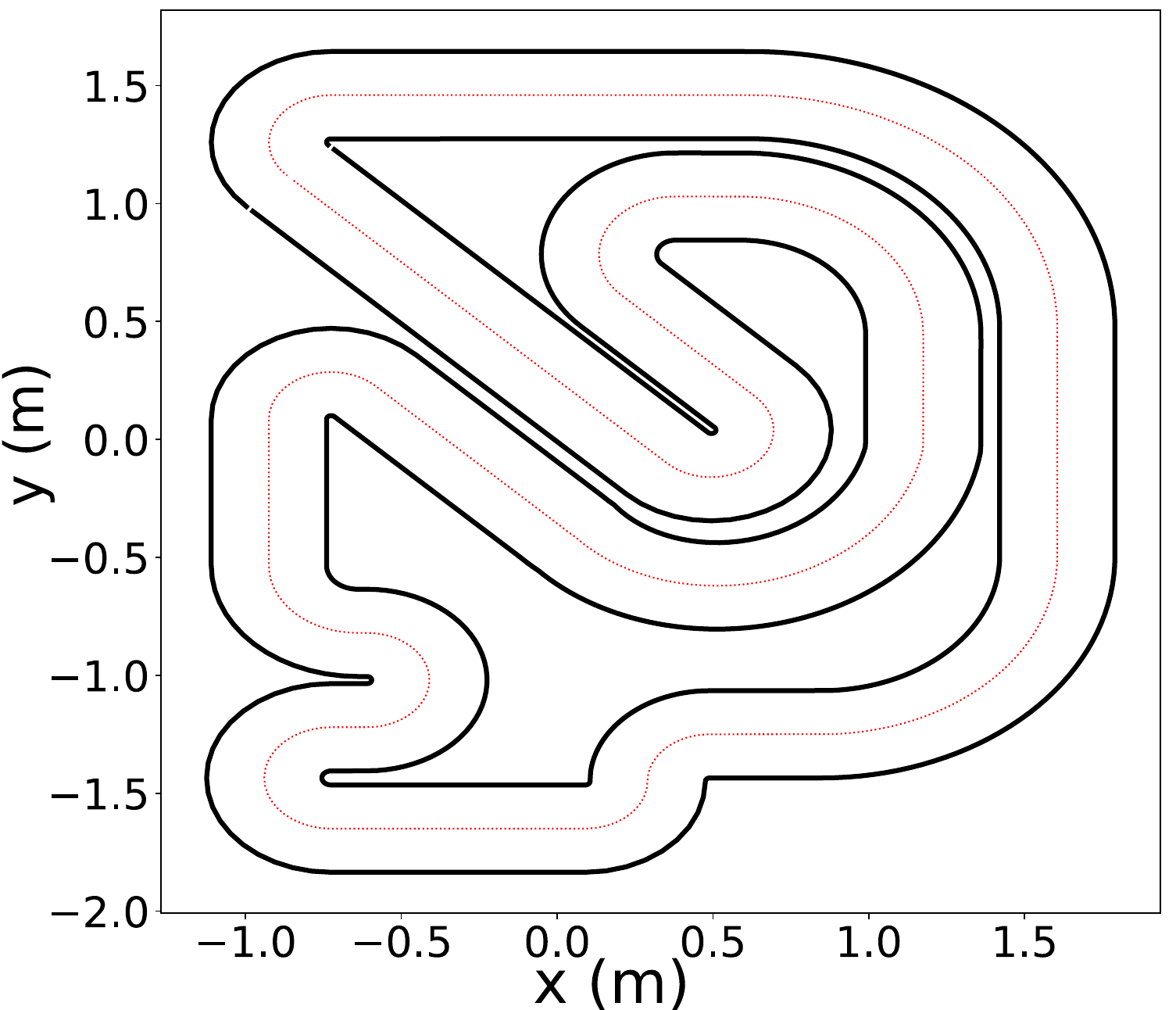}
	\caption{The racetrack utilized for data collection.}
	\label{fig:ETHZTrack}
\end{figure}

\section{Result and Discussion}
\label{sec:result}
\subsection{Training Dataset}
We compare the simulation performance of the proposed FTHD against the state-of-the-art DDM, using data collected from the simulator described in~\cite{JainRaceOpt2020}. This simulator features a 1:43 scaled racecar engaged in pure-pursuit of a reference trajectory on the racetrack depicted in Fig.~\ref{fig:ETHZTrack}. The ground truth (GT) of the scaled racecar's coefficients is available but only used for comparison after training. The sampling frequency is 50 Hz, resulting in a time interval between samples \(\Delta t_\text{sim}\) is 0.02 s. The total size of the simulation dataset \(\mathbb{D}^\text{sim}_\text{total}\) consists of 1000 samples.

\renewcommand{\arraystretch}{1.1}
\begin{table*}[!t]
	\caption{Coefficient ranges for the Dynamic STM}
	\label{tab:coefficient_range}
	\vspace{-3mm}
	% \centering
	\begin{tabular*}{\columnwidth}{@{\extracolsep{\fill}}c|c|c|c|c|c|c|c}
		\hline
		\hline
		\multicolumn{1}{c|}{\multirow{2}{*}{Layers}} & \multicolumn{1}{c|}{\multirow{2}{*}{Coefficient}} & \multicolumn{2}{c|}{Sim \([\underaccent{\rule{.4em}{.8pt}}{\Phi}_u,\accentset{\rule{.4em}{.8pt}}{\Phi}_u]\)} & \multicolumn{2}{c|}{Real \([\underaccent{\rule{.4em}{.8pt}}{\Phi}_u,\accentset{\rule{.4em}{.8pt}}{\Phi}_u]\)} & \multicolumn{2}{c}{Real \([\underaccent{\rule{.4em}{.8pt}}{\Phi}_u^*,\accentset{\rule{.4em}{.8pt}}{\Phi}_u^*]\)} \\ \cline{3-8}
		\multicolumn{1}{c|}{} & \multicolumn{1}{c|}{}  &\multicolumn{1}{c}{Front} & \multicolumn{1}{c|}{Rear} & \multicolumn{1}{c}{Front}      &\multicolumn{1}{c|}{Rear} & \multicolumn{1}{c}{Front}      &\multicolumn{1}{c}{Rear}\\ \hline %\hline
		\multicolumn{1}{c|}{\multirow{6}{*}{Pacejka Layer}} & \multicolumn{1}{c|}{$B$} & \multicolumn{1}{c}{\([5.0,30.0]\)} & \multicolumn{1}{c|}{\([5.0,30.0]\)} & \multicolumn{1}{c}{\([5.0,30.0]\)} & \multicolumn{1}{c|}{\([5.0,30.0]\)} & \multicolumn{1}{c}{\([1.0,20.0]\)} & \multicolumn{1}{c}{\([1.0,20.0]\)} \\
		\multicolumn{1}{c|}{} & \multicolumn{1}{c|}{$C$} & \multicolumn{1}{c}{\([0.5,2.0]\)} & \multicolumn{1}{c|}{\([0.5,2.0]\)} & \multicolumn{1}{c}{\([0.5,2.0]\)} & \multicolumn{1}{c|}{\([0.5,2.0]\)} & \multicolumn{1}{c}{\([0.1,1.5]\)} & \multicolumn{1}{c}{\([0.1,1.5]\)} \\ 
		\multicolumn{1}{c|}{} & \multicolumn{1}{c|}{$D$} & \multicolumn{1}{c}{\([0.1,1.9]\)} & \multicolumn{1}{c|}{\([0.1,1.9]\)} & \multicolumn{1}{c}{\([100,10^{4}]\)} & \multicolumn{1}{c|}{\([100,10^{4}]\)} & \multicolumn{1}{c}{\([10,8\times 10^{3}]\)} & \multicolumn{1}{c}{\([10,8\times 10^{3}]\)} \\ 
		\multicolumn{1}{c|}{} & \multicolumn{1}{c|}{$E$} & \multicolumn{1}{c}{\([-2.0,0.0]\)} & \multicolumn{1}{c|}{\([-2.0,0.0]\)} & \multicolumn{1}{c}{\([-2.0,0.0]\)} & \multicolumn{1}{c|}{\([-2.0,0.0]\)} & \multicolumn{1}{c}{\([-2.0,5.0]\)} & \multicolumn{1}{c}{\([-2.0,10.0]\)} \\ 
		\multicolumn{1}{c|}{} & \multicolumn{1}{c|}{\(S_v\)} & \multicolumn{1}{c}{\([-0.003,0.003]\)} & \multicolumn{1}{c|}{\([-0.003,0.003]\)} & \multicolumn{1}{c}{\([-300,300]\)} & \multicolumn{1}{c|}{\([-300,300]\)} & \multicolumn{1}{c}{\([-2\times 10^{3},300]\)} & \multicolumn{1}{c}{\([-2\times 10^{3},300]\)} \\ 
		\multicolumn{1}{c|}{} & \multicolumn{1}{c|}{\(S_h\)} & \multicolumn{1}{c}{\([-0.02,0.02]\)} & \multicolumn{1}{c|}{\([-0.02,0.02]\)} & \multicolumn{1}{c}{\([-0.02,0.02]\)} & \multicolumn{1}{c|}{\([-0.02,0.02]\)} & \multicolumn{1}{c}{\([-0.02,0.2]\)} & \multicolumn{1}{c}{\([-0.02,0.2]\)} \\ \hline 
		%\hline
		\multicolumn{1}{c|}{\multirow{4}{*}{Drivetrain Layer}} & \multicolumn{1}{c|}{\(C_{m1}\) (N)} & \multicolumn{2}{c|}{\([0.1435,0.574]\)}  & \multicolumn{2}{c|}{\([500,2\times 10^{3}]\)} & \multicolumn{2}{c}{\([100,1\times 10^{4}]\)} \\ 
		\multicolumn{1}{c|}{} & \multicolumn{1}{c|}{\(C_{m2}\) (kg/s)} & \multicolumn{2}{c|}{\([0.0273,0.109]\)}  & \multicolumn{2}{c|}{\([0,1.0]\)} & \multicolumn{2}{c}{\([0,5.0]\)} \\ 
		\multicolumn{1}{c|}{} & \multicolumn{1}{c|}{\(C_{r0}\) (N)} & \multicolumn{2}{c|}{\([0.0259,0.1036]\)}  & \multicolumn{2}{c|}{\([0.1,1.4]\)} & \multicolumn{2}{c}{\([0.1,1.4]\)} \\ 
		\multicolumn{1}{c|}{} & \multicolumn{1}{c|}{\(C_{d}\) (kg/m)} & \multicolumn{2}{c|}{\([1.75\times 10^{-4},0.1036]\)}  & \multicolumn{2}{c|}{\([0.1,1.4]\)} & \multicolumn{2}{c}{\([0.1,1.4]\)} \\ \hline
		%\hline
		\multicolumn{1}{c|}{Inertia Layer} & \multicolumn{1}{c|}{\(I_{z}\) (kg$\cdot \text{m}^{2}$)} & \multicolumn{2}{c|}{\([1.39\times 10^{-5},5.56\times 10^{-5}]\)}  & \multicolumn{2}{c|}{\([5\times 10^{5},2\times 10^{3}]\)} & \multicolumn{2}{c}{\([5\times 10^{3},2\times 10^{4}]\)} \\ \hline\hline
	\end{tabular*}
\end{table*}

To further evaluate the real-world performance of the FTHD model and compare it with DDM in terms of the accuracy of the filtered data and adjusted parameter ranges derived from EKF-FTHD, we use a real-world dataset from a full-scale Indy autonomous racecar~\cite{kulkarni2023racecar}. The dataset includes fused measurements from two RTK-corrected GNSS signals and an IMU, which are passed through a 100 Hz EKF. The sampling frequency of the real data is 25 Hz, resulting in a time interval between samples of \(\Delta t_\text{exp} = \) 0.04 s. The total dataset size, denoted as \(\mathbb{D}^\text{exp}_\text{total}\), consists of 11,488 samples collected at the Putnam Park Road Course in Indianapolis. The EKF-FTHD filters the full dataset, \(\mathbb{D}^\text{exp}_\text{total}\), and output a new dataset, denoted as \(\mathbb{D}^\text{EKF}\), which is then used to train both the FTHD and DDM models for prediction. 

To demonstrate the robustness of our method compared to DDM when reducing the training dataset size, we simulate random selections of \(30\%\), \(20\%\) and \(15\%\) of the dataset samples \(\mathbb{D}^\text{sim}_\text{train}\) for training, with \(\mathbb{D}^\text{sim}_\text{total}\) used for validation. For real-world experiments, we randomly select \(90\%\), \(60\%\), \(30\%\), \(15\%\) and \(5\%\) of the dataset samples for training, denoted as \(\mathbb{D}^\text{exp}_\text{train}\) and \(\mathbb{D}^\text{EKF}_\text{train}\). The full datasets \(\mathbb{D}^\text{exp}_\text{total}\) and \(\mathbb{D}^\text{EKF}\) are used for validation. Both models undergo the same number of training iterations, using identical hyperparameter tuning to minimize validation loss.

\begin{figure*}[t!]
	%\subfigure[]{
		%	\hspace{-5mm}
		% \label{fig:ForceResponse80}
		%	\includegraphics[width=0.34\linewidth]{photo/poseter_trained_force_benchmark_80ins.pdf}}
	%\subfigure[]{
		%	\hspace{-5mm}
		% \label{fig:ForceResponse50}
		%	\includegraphics[width=0.33\linewidth]{photo/poseter_trained_force_benchmark_50ins2.pdf}}
	\subfigure[]{
		\hspace{-2mm}
		\label{fig:ForceResponse30}
		\includegraphics[width=0.33\linewidth]{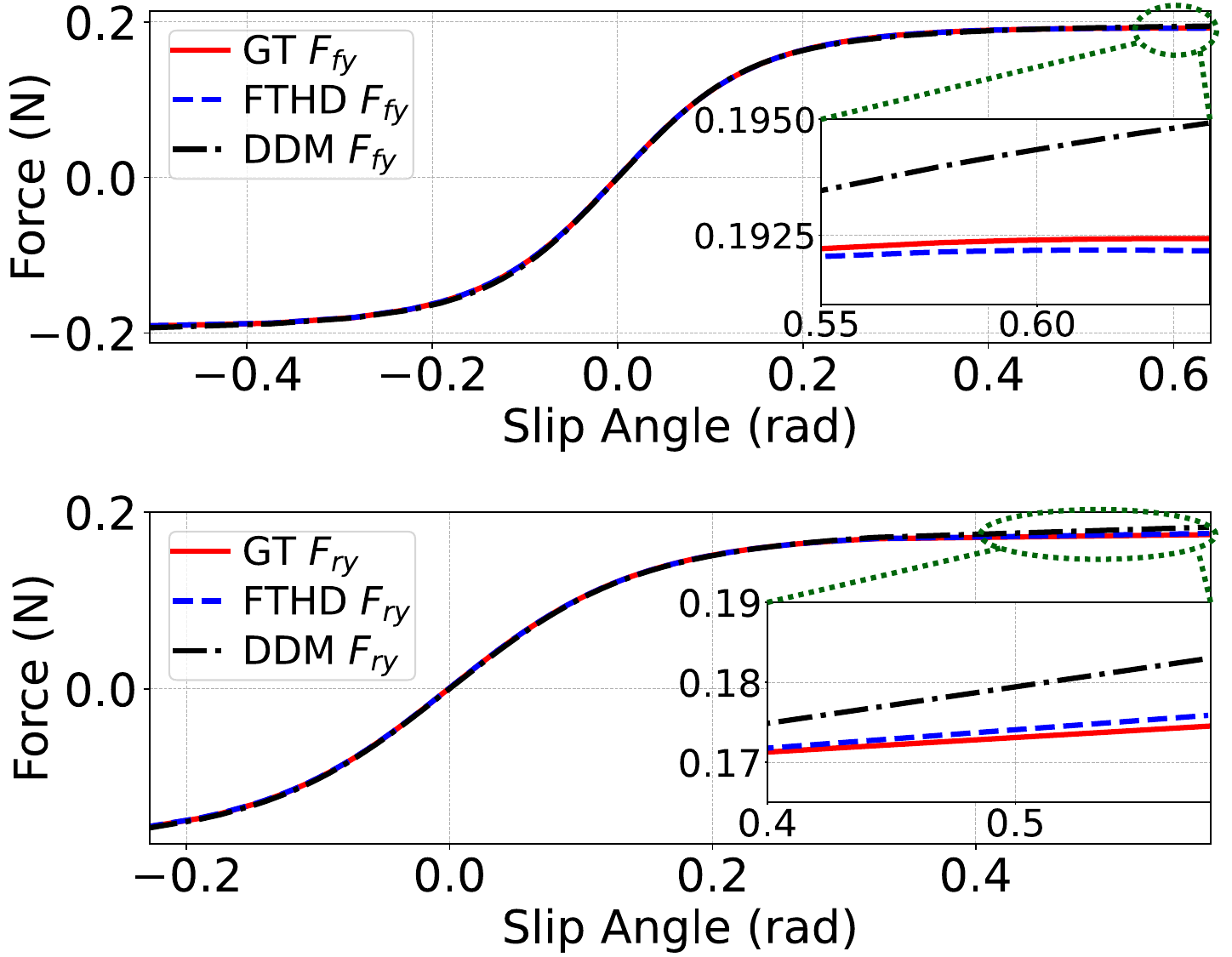}}
	\subfigure[]{\hspace{-2mm}
		\label{fig:ForceResponse20}
		\includegraphics[width=0.33\linewidth]{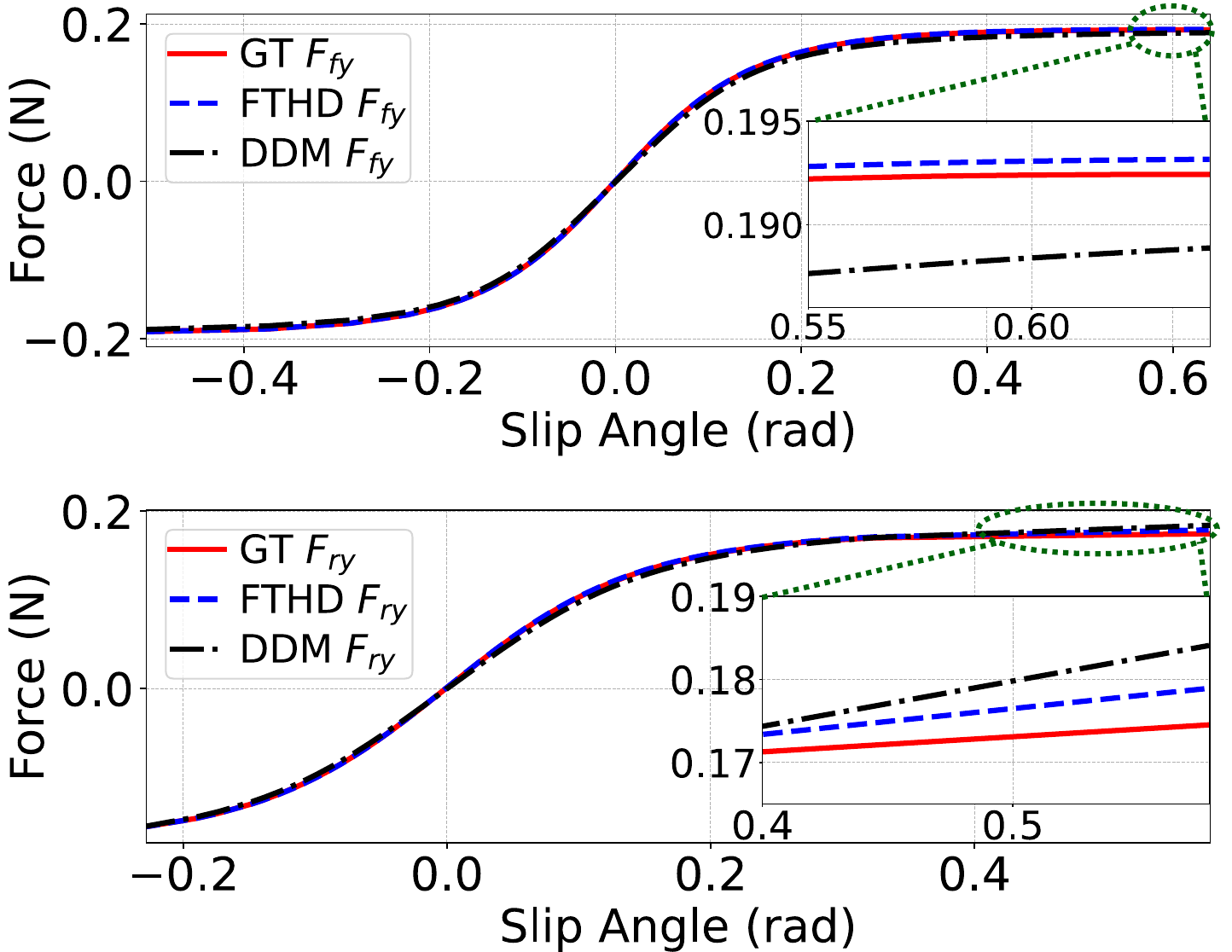}}
	\subfigure[]{\hspace{-2mm}
		\label{fig:ForceResponse15}
		\includegraphics[width=0.325\linewidth]{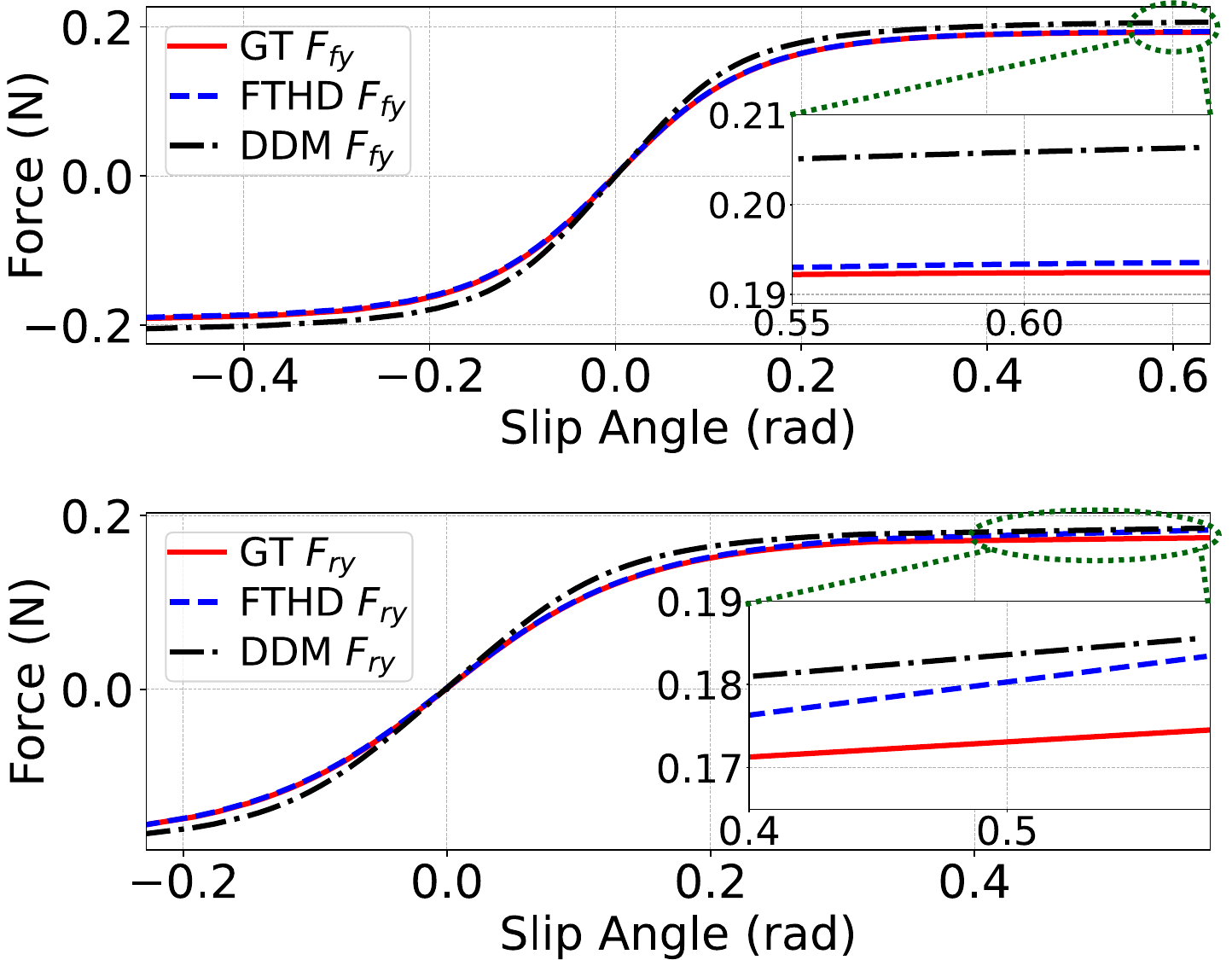}}
	\caption{Comparison of the front wheel lateral force $F_{fy}$ (top) and the rear wheel lateral force $F_{ry}$ (bottom) using the trained coefficients with the proposed FTHD scheme, DDM model, and the GT. The force responses of the trained coefficients using (a) 30\%, (b) 20\%, and (c) 15\% of the total training set. (inset) Zoomed-in comparison.} %The '*' markers indicates the nanowires initial positions, and the '$\circ$' markers indicates the targeting areas.} 
\label{fig:ForceResponse}
\end{figure*}

\begin{figure*}[t!]
	\centering
	% \subfigure[]{
		% 	\hspace{-2mm}
		% 	\label{fig:AD80}
		% 	\includegraphics[width=0.33\linewidth]{photo/simulated_data_analysis_80.pdf}}
	% \subfigure[]{
		% 	\hspace{-3mm}
		% 	\label{fig:AD50}
		% 	\includegraphics[width=0.33\linewidth]{photo/simulated_data_analysis_50.pdf}}
	\subfigure[]{
		\hspace{-2mm}
		\label{fig:AD30}
		\includegraphics[width=0.33\linewidth]{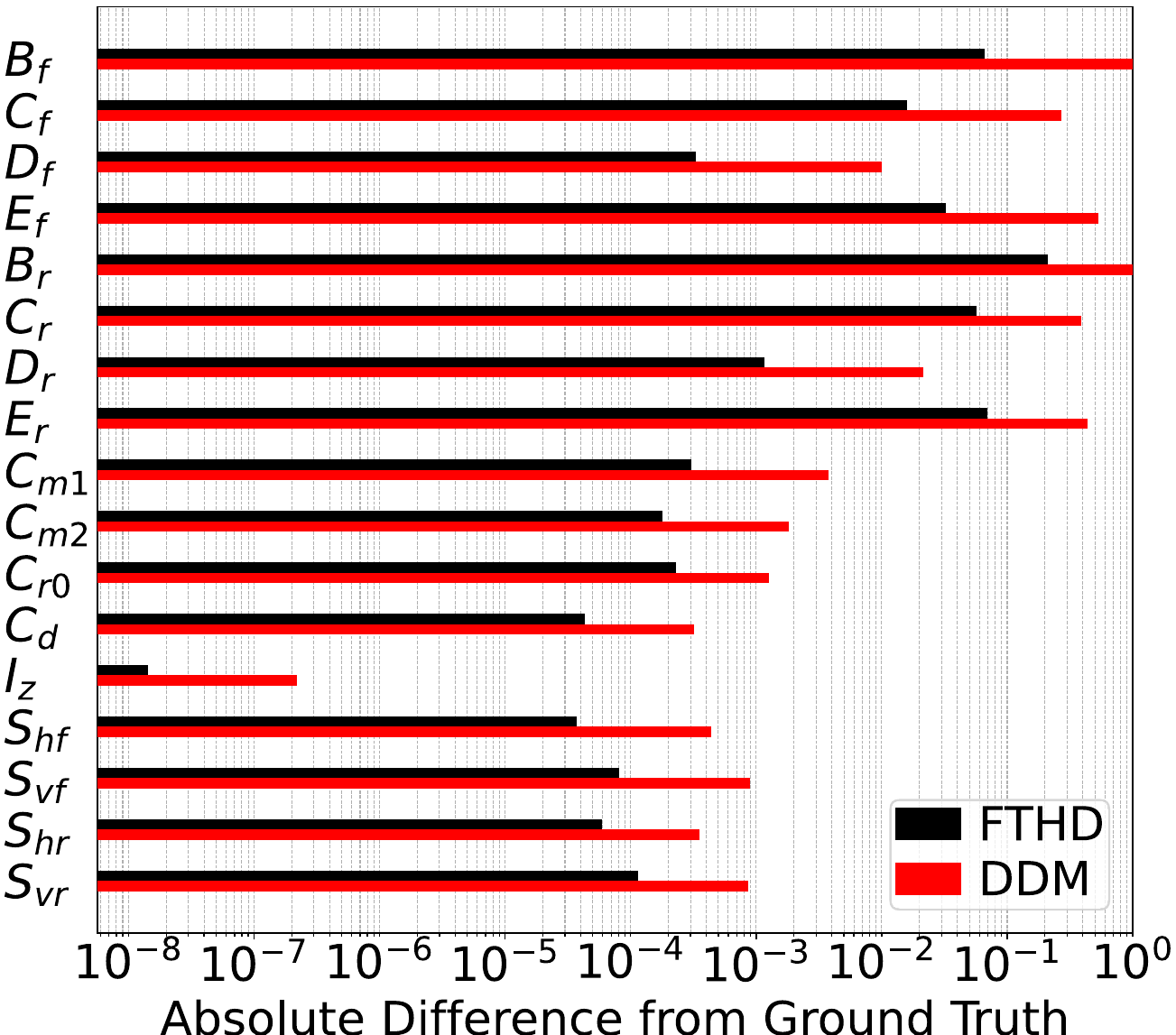}}
	\subfigure[]{\hspace{-3mm}
		\label{fig:AD20}
		\includegraphics[width=0.33\linewidth]{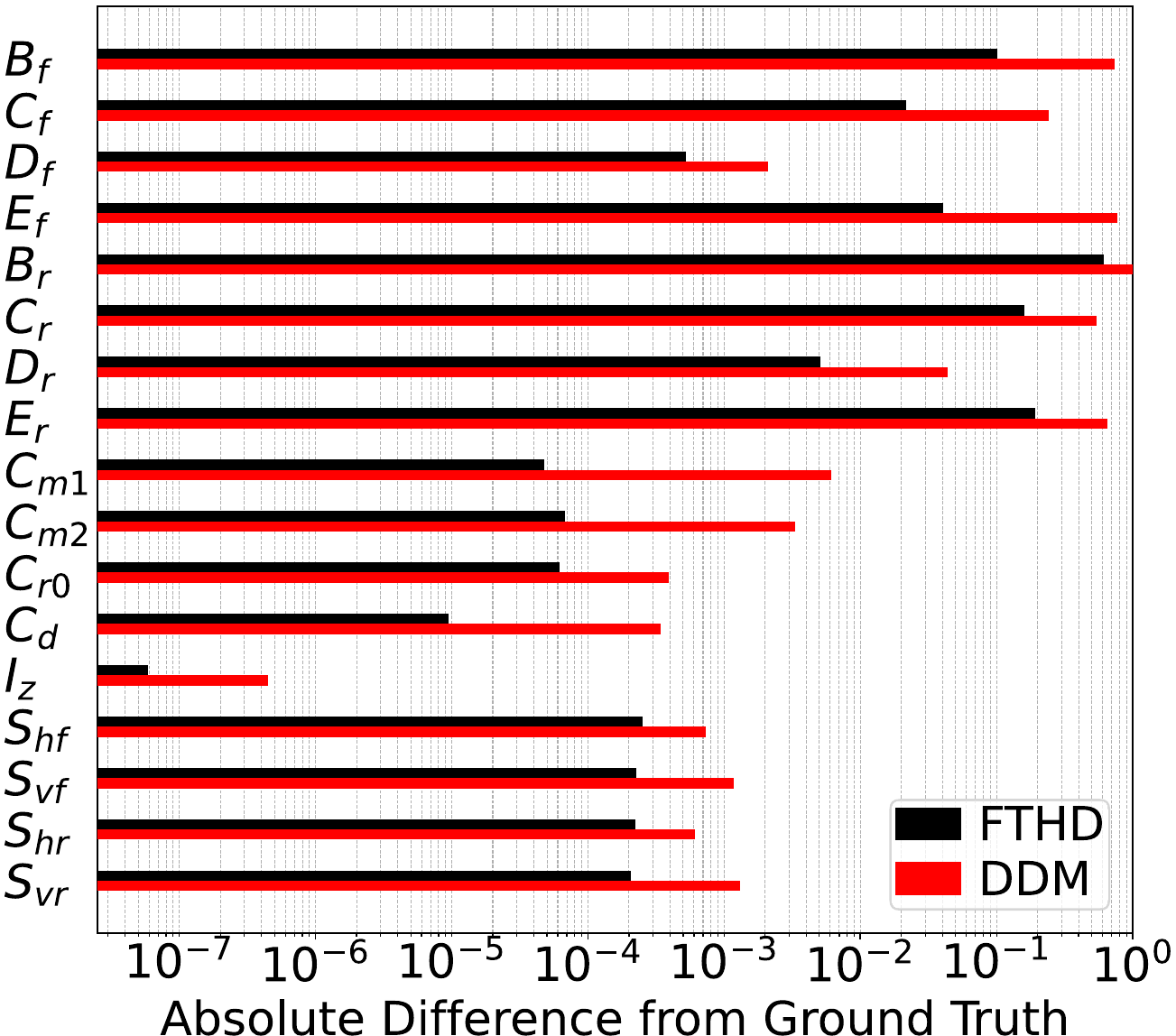}}
	\subfigure[]{\hspace{-3mm}
		\label{fig:AD15}
		\includegraphics[width=0.33\linewidth]{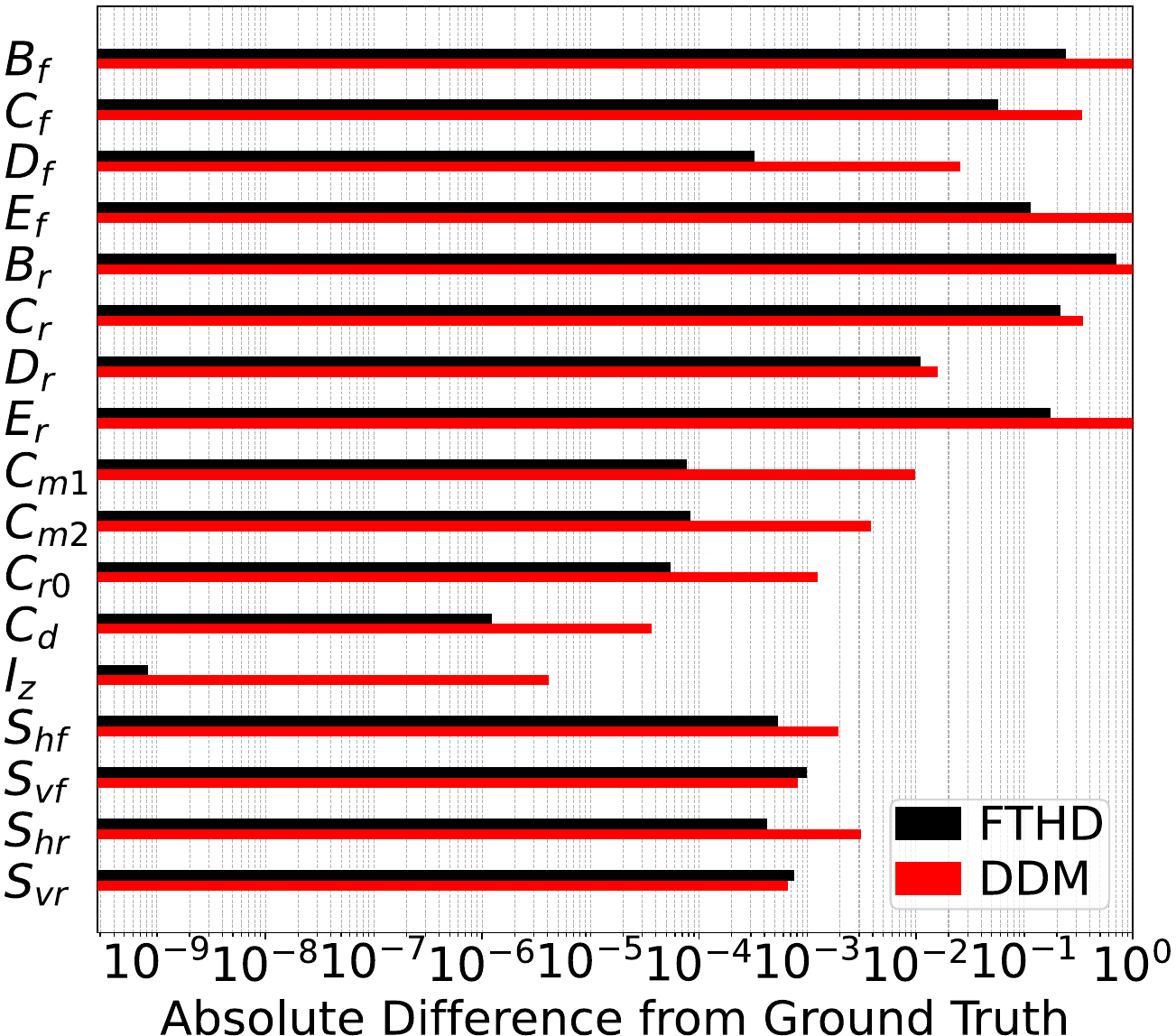}}
	\caption{Absolute difference between FTHD and DDM estimated Pacejka, Drivetrain and moment of inertia coefficients compared to the GT using (a) 30\%, (b) 20\%, and (c) 15\% of the total training set.} %(a) 80\%, (b) 50\%,  (c) 30\%, (d) 20\%, and (e) 15\% of the total training set.} %The '*' markers indicates the nanowires initial positions, and the '$\circ$' markers indicates the targeting areas.} 
	\label{fig:AD}
\end{figure*}

\subsection{Model Coefficient Ranges}
To estimate the physics coefficient values \(\hat{\Phi}_u\) in the physics guard layer of the model, as shown in Fig.~\ref{fig:EKF_Fine-tuningModel}, as well as the estimated EKF covariance values \(\hat{\Phi}_f\), the bounds \(\underaccent{\rule{.4em}{.8pt}}{\Phi}_u \) and \(\accentset{\rule{.4em}{.8pt}}{\Phi}_u\), as well as \(\underaccent{\rule{.4em}{.8pt}}{\Phi}_f\) and \(\accentset{\rule{.4em}{.8pt}}{\Phi}_f\), are required. For simulation experiments, we use the known GT to compare the performance of FTHD with DDM, using the same coefficient ranges as in~\cite{chrosniak2023deep}. In real-world experiments, the bounds \([\underaccent{\rule{.4em}{.8pt}}{\Phi}_u , \accentset{\rule{.4em}{.8pt}}{\Phi}_u]\) used in~\cite{chrosniak2023deep} are adjusted to \([\underaccent{\rule{.4em}{.8pt}}{\Phi}_u^* , \accentset{\rule{.4em}{.8pt}}{\Phi}_u^*]\) based on EKF-FTHD output. The covariance bounds \([\underaccent{\rule{.4em}{.8pt}}{\Phi}_f ,\accentset{\rule{.4em}{.8pt}}{\Phi}_f]\) are set according to the quality of the raw data, where better-quality data lead to smaller covariance values. These ranges are detailed in Table~\ref{tab:coefficient_range}, and Table~\ref{tab:ekf_covariance_range} provides the EKF covariance matrix ranges.

\renewcommand{\arraystretch}{1.1}
\begin{table}[!h]
	\caption{Coefficient ranges of the covariance matrices for the EKF-FTHD model}
	\label{tab:ekf_covariance_range}
	\vspace{-3mm}
	\centering
	\begin{tabular*}{\columnwidth}{@{\extracolsep{\fill}}c|c|c|c}
		\hline
		\hline
		\multicolumn{1}{c|}{\multirow{2}{*}{Layer}} & \multicolumn{1}{c}{\multirow{2}{*}{Coefficient}} & \multicolumn{2}{|c}{Real Data} \\ \cline{3-4}
		\multicolumn{1}{c|}{} & \multicolumn{1}{c}{}  &\multicolumn{1}{|c}{Min} &\multicolumn{1}{c}{Max} \\ \hline
		\multicolumn{1}{c|}{\multirow{6}{*}{\text{EKF} Layer}} & \multicolumn{1}{c}{\(q_{v_x}\)} & \multicolumn{1}{|c}{0.1} & \multicolumn{1}{c}{1} \\ 
		\multicolumn{1}{c|}{} & \multicolumn{1}{c}{\(q_{v_y}\)} & \multicolumn{1}{|c}{0.1} & \multicolumn{1}{c}{1} \\ 
		\multicolumn{1}{c|}{} & \multicolumn{1}{c}{\(q_{\omega}\)} & \multicolumn{1}{|c}{0.1} & \multicolumn{1}{c}{1} \\ 
		\multicolumn{1}{c|}{} & \multicolumn{1}{c}{\(r_{v_x}\)} & \multicolumn{1}{|c}{0.01} & \multicolumn{1}{c}{1.0} \\
		\multicolumn{1}{c|}{} & \multicolumn{1}{c}{\(r_{v_y}\)} & \multicolumn{1}{|c}{0.01} & \multicolumn{1}{c}{1.0} \\
		\multicolumn{1}{c|}{} & \multicolumn{1}{c}{\(r_{\omega}\)} & \multicolumn{1}{|c}{\(1\times 10^{-4}\)} & \multicolumn{1}{c}{\(0.01\)} \\ \hline\hline
	\end{tabular*}
\end{table}
\begin{figure}[ht!]
	\centering
	\includegraphics[width=\linewidth]{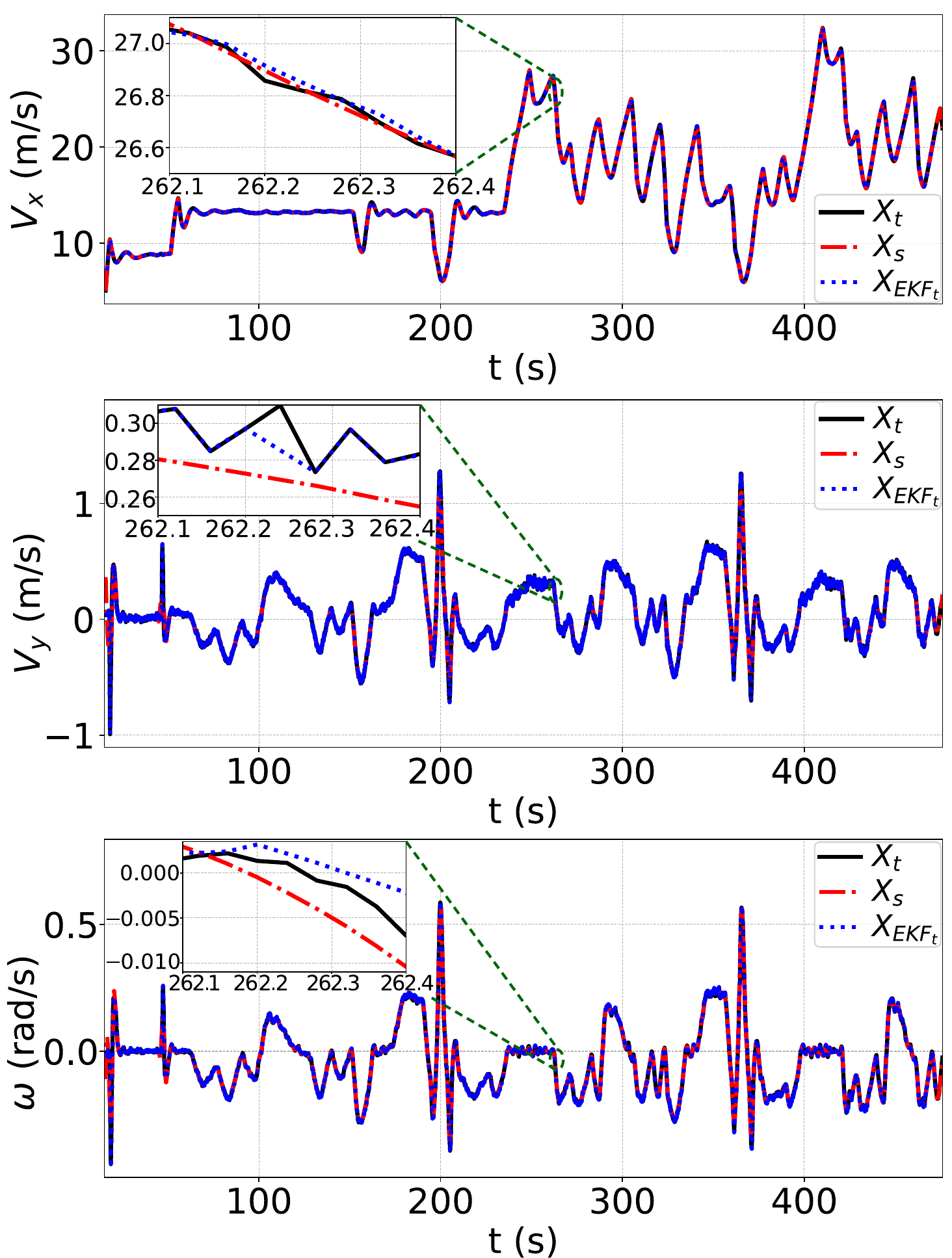}
	\caption{Comparison of raw data, traditionally smoothed data, and EKF-filtered data in the real-world dataset. (inset) Zoomed-in comparison.}
	\label{fig:smooth data compare}
\end{figure}

\subsection{Evaluation Metrics}
For the simulation, with known GT, we use the Pacejka and drivetrain coefficients that result in the lowest validation loss from both FTHD and DDM to calculate lateral forces using the same STM. During the fine-tuning hybrid training of FTHD, we choose weight parameters $w_1=0.99975$ and $w_2=0.00025$. The comparison results with the GT are shown in Fig.~\ref{fig:ForceResponse}. It's evident that as the size of the training set decreases, the FTHD retains better alignment with the GT outputs compared to DDM. Furthermore, it's clear to see that when only 15\% of the training set is used, the force plots of DDM show much greater deviation from the GT, while FTHD can still closely approximate it.

To evaluate the estimation performance of FTHD and DDM in simulation, we compare the minimum validation loss, $L_\text{min}=\min\left(Loss_\text{val}\right)$, the root-mean-square-error (RMSE) of velocities, and the maximum errors of velocities (\(\epsilon_{max}\)) for \(v_x\), \(v_y\) and \(\omega\), as used in~\cite{chrosniak2023deep}. Additionally, we evaluate the absolute difference between the parameters estimated by FTHD and DDM from the GT, using datasets reduced to \(30\%\), \(20\%\) and \(15\%\) of the total sample data. The results, shown in Table~\ref{tab:RMSEValLossMaxE} and Fig.~\ref{fig:AD} reveal that even with the same number of iterations (15k in total) for both models during training, the FTHD outperforms the supervised-only PINN. The estimated Pacejka coefficients also exhibit less difference with the GT. Additionally, even when the dataset is reduced, the FTHD maintains significantly higher accuracy in estimating the moment of inertia \(I_z\) compared to DDM.

 \renewcommand{\arraystretch}{1.1}
 \begin{table}[th!]
 	\caption{Validation loss comparison of DDM trained on original dataset and EKF dataset, with preset and EKF-FTHD adjusted parameter ranges, across varying training data ratios}
 	%Validation results of DDM while using \(\mathbb{D}^\text{exp}_\text{total}\) and \(\mathbb{D}^\text{EKF}\) while the training dataset decreases}
 \label{tab:DDM with different dataset and ranges}
 \begin{center}
 	\vspace{-3mm}
 	\begin{tabular}{c|c|c}
 		\hline
 		\hline
 		\(\frac{\mathbb{D}^\text{exp}_\text{train}}{\mathbb{D}^\text{exp}_\text{total}}\) or \(\frac{\mathbb{D}^\text{EKF}_\text{train}}{\mathbb{D}^\text{EKF}}\) & \(\mathbb{D}^\text{exp}_\text{total}\) with \([\underaccent{\rule{.4em}{.8pt}}{\Phi}_u , \accentset{\rule{.4em}{.8pt}}{\Phi}_u]\) & \(\mathbb{D}^\text{EKF}\) with \([\underaccent{\rule{.4em}{.8pt}}{\Phi}_u^* , \accentset{\rule{.4em}{.8pt}}{\Phi}_u^*]\)  \\\hline 
 		\(90\%\) & 1.480$\times 10^{-4}$ & 7.004$\times 10^{-5}$  \\\hline
 		\(60\%\) & 2.132$\times 10^{-4}$ & 1.235$\times 10^{-4}$  \\\hline
 		\(30\%\) & 2.922$\times 10^{-4}$ & 1.894$\times 10^{-4}$  \\\hline
 		\(15\%\) & 3.290$\times 10^{-4}$ & 2.220$\times 10^{-4}$  \\\hline
 		\(5\%\) & 3.859$\times 10^{-4}$ & 2.854$\times 10^{-4}$  \\\hline\hline
 		% DDM use \(\mathbb{D}^\text{EKF}\) and \([\underaccent{\rule{.4em}{.8pt}}{\Phi}_u^* , \accentset{\rule{.4em}{.8pt}}{\Phi}_u^*]\) & 0.000159 \\\hline
 		% FTHD use \(\mathbb{D}^\text{EKF}\) and \([\underaccent{\rule{.4em}{.8pt}}{\Phi}_u^* , \accentset{\rule{.4em}{.8pt}}{\Phi}_u^*]\) & 0.0000425 \\\hline
 	\end{tabular}
 \end{center}
 \end{table}
 
In the real-world experiments, we use \(\mathbb{D}^\text{EKF}\) for training the FTHD model. Fig.~\ref{fig:smooth data compare} presents the plots of \(v_x\), \(v_y\) and \(\omega\) over time, where \(X_t\) refers to the raw data from \(\mathbb{D}^\text{exp}_\text{total}\), \(X_s\) represents the traditionally smoothed data, and \(X_{\text{\text{EKF}}_t}\) corresponds to the data from \(\mathbb{D}^\text{EKF}\). Upon comparison, it is clear that although \(X_s\) smooths the data, it results in a loss of key features and alters the relationship between the states and controls. In contrast, \(X_{\text{\text{EKF}}_t}\) separates noise from \(X_t\) while preserving the physically meaningful aspects of the data.

\begin{figure}[t!]
	\centering
	\includegraphics[width=\linewidth]{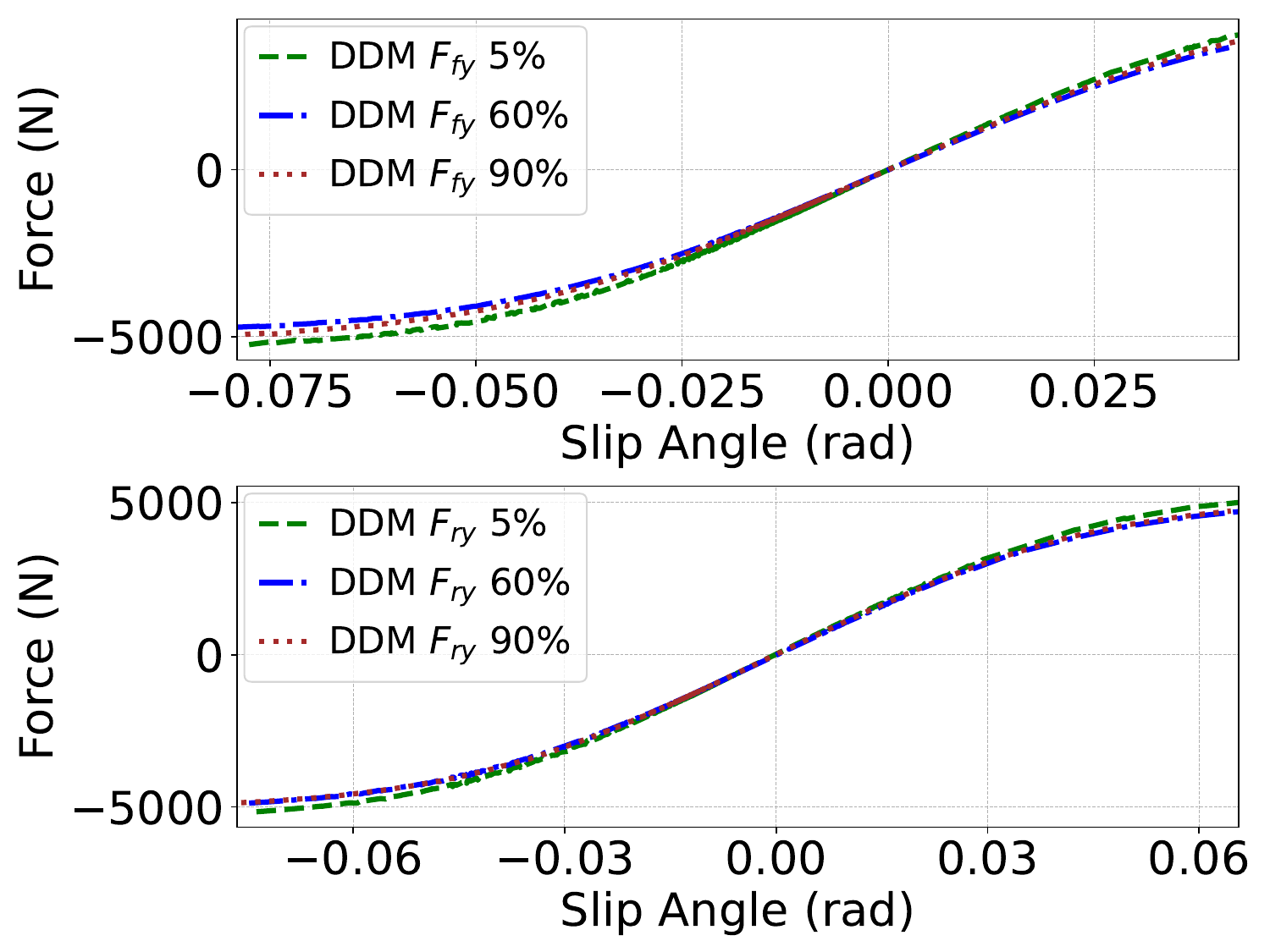}
	\caption{Lateral force response of the DDM model in real-world experiments as the size of the training set decreases.}
	\label{fig:Real_force_DDM}
\end{figure}

\begin{figure}[t!]
	\centering
	\includegraphics[width=\linewidth]{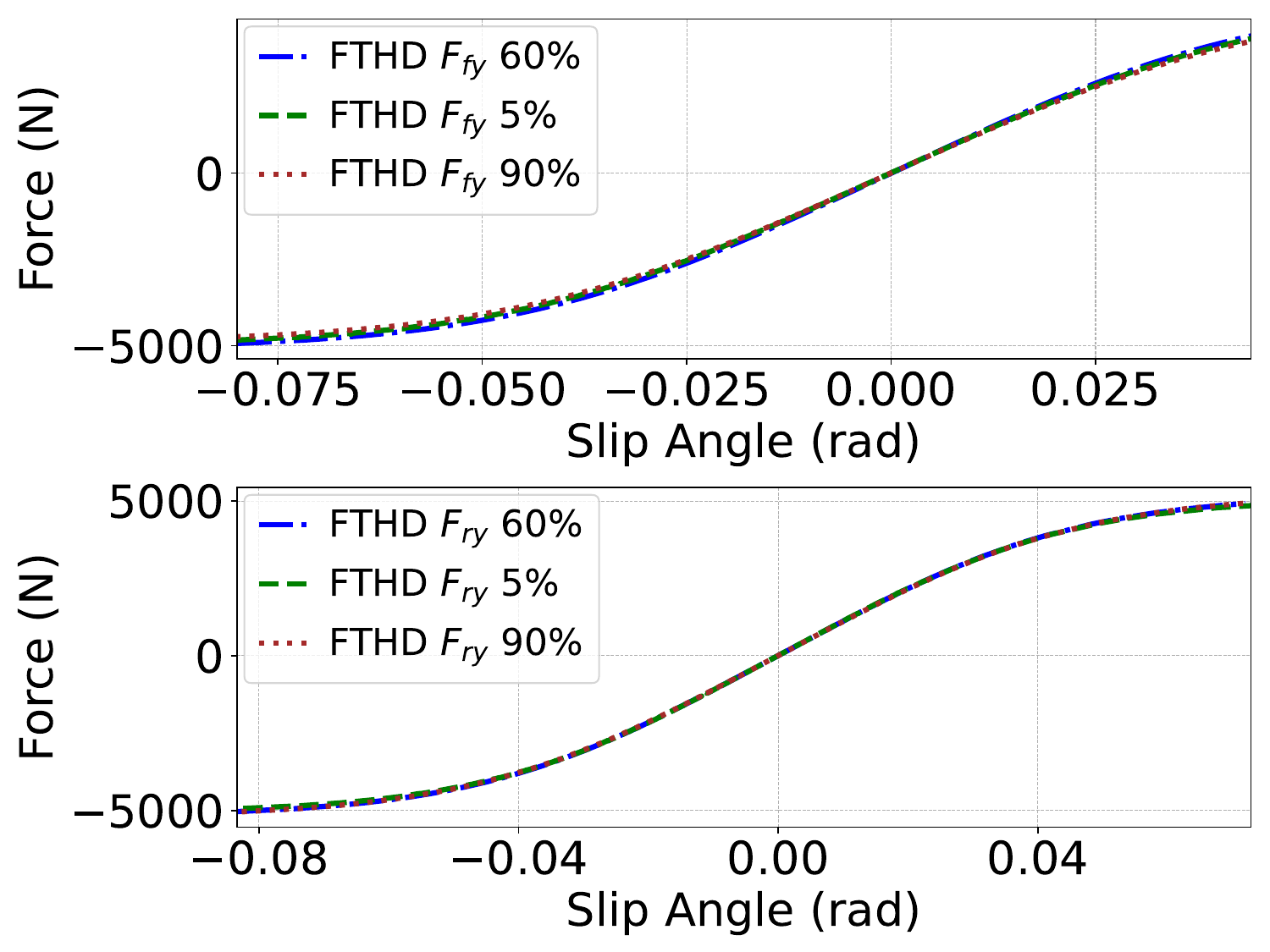}
	\caption{Lateral force response of the FTHD model in real-world experiments as the size of the training set decreases.}
	\label{fig:Real_force_FTHD}
\end{figure}

To demonstrate the improvement in data fidelity after the EKF-FTHD processing, we train DDM models on both \(\mathbb{D}^\text{exp}_\text{total}\) and \(\mathbb{D}^\text{EKF}\), with preset parameter ranges \([\underaccent{\rule{.4em}{.8pt}}{\Phi}_u , \accentset{\rule{.4em}{.8pt}}{\Phi}_u]\) and EKF-FTHD adjusted ranges \([\underaccent{\rule{.4em}{.8pt}}{\Phi}_u^* , \accentset{\rule{.4em}{.8pt}}{\Phi}_u^*]\). The validation losses, presented in Table~\ref{tab:DDM with different dataset and ranges}, show that with the same structured model, the filtered dataset \(\mathbb{D}^\text{EKF}\) achieves higher accuracy across different training data ratios.

\begin{figure}[th!]
	\centering
	\includegraphics[width=\linewidth]{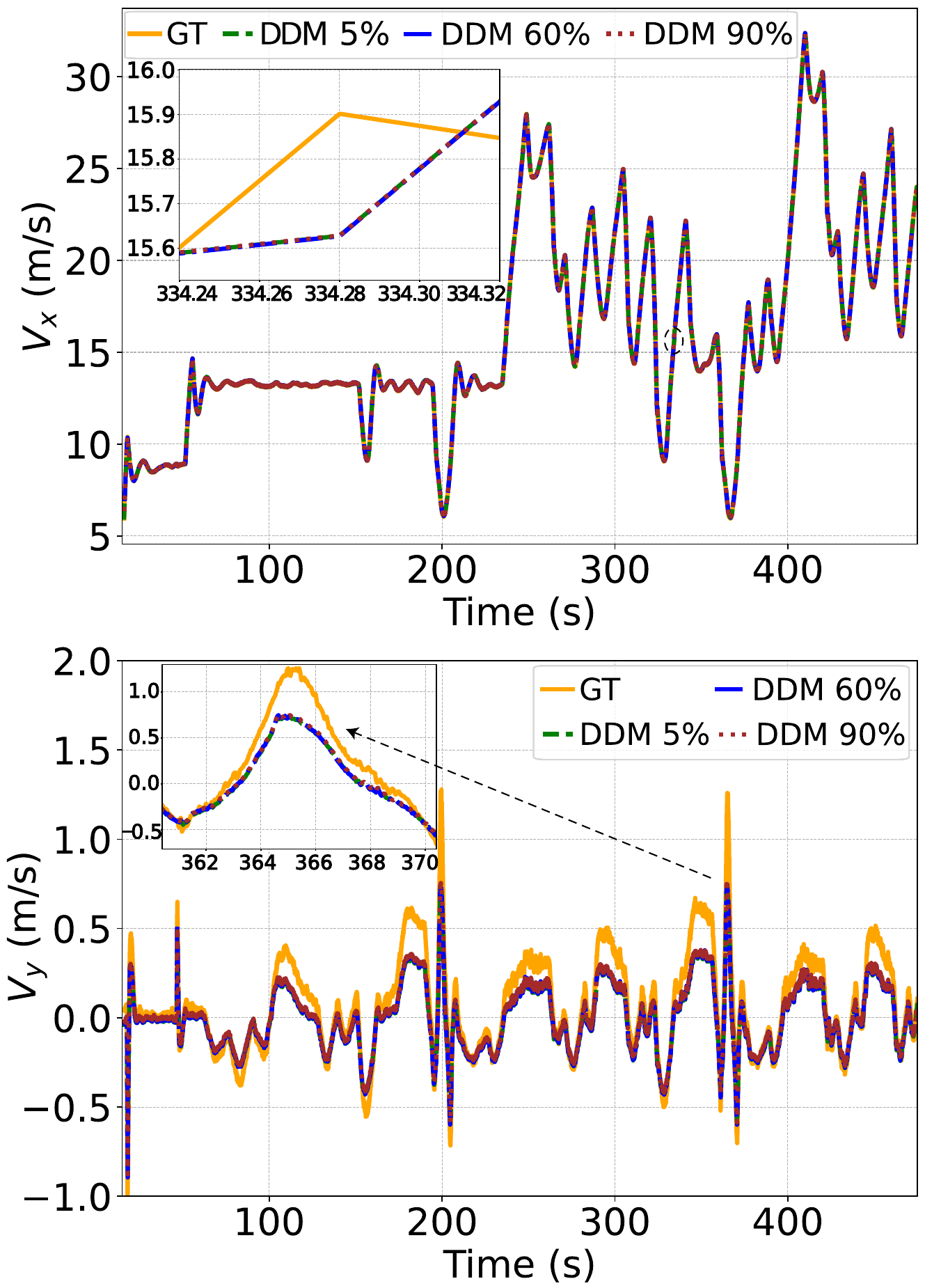}
	\caption{Velocity prediction of DDM in real experiments as the training set size decreases. (inset) The zoomed-in view highlights the maximum error compared to the GT at the time marked using black dashed circles or lines in the figures.}
	\label{fig:Real_velocity_DDM}
\end{figure}

\begin{figure}[th!]
	\centering
	\includegraphics[width=\linewidth]{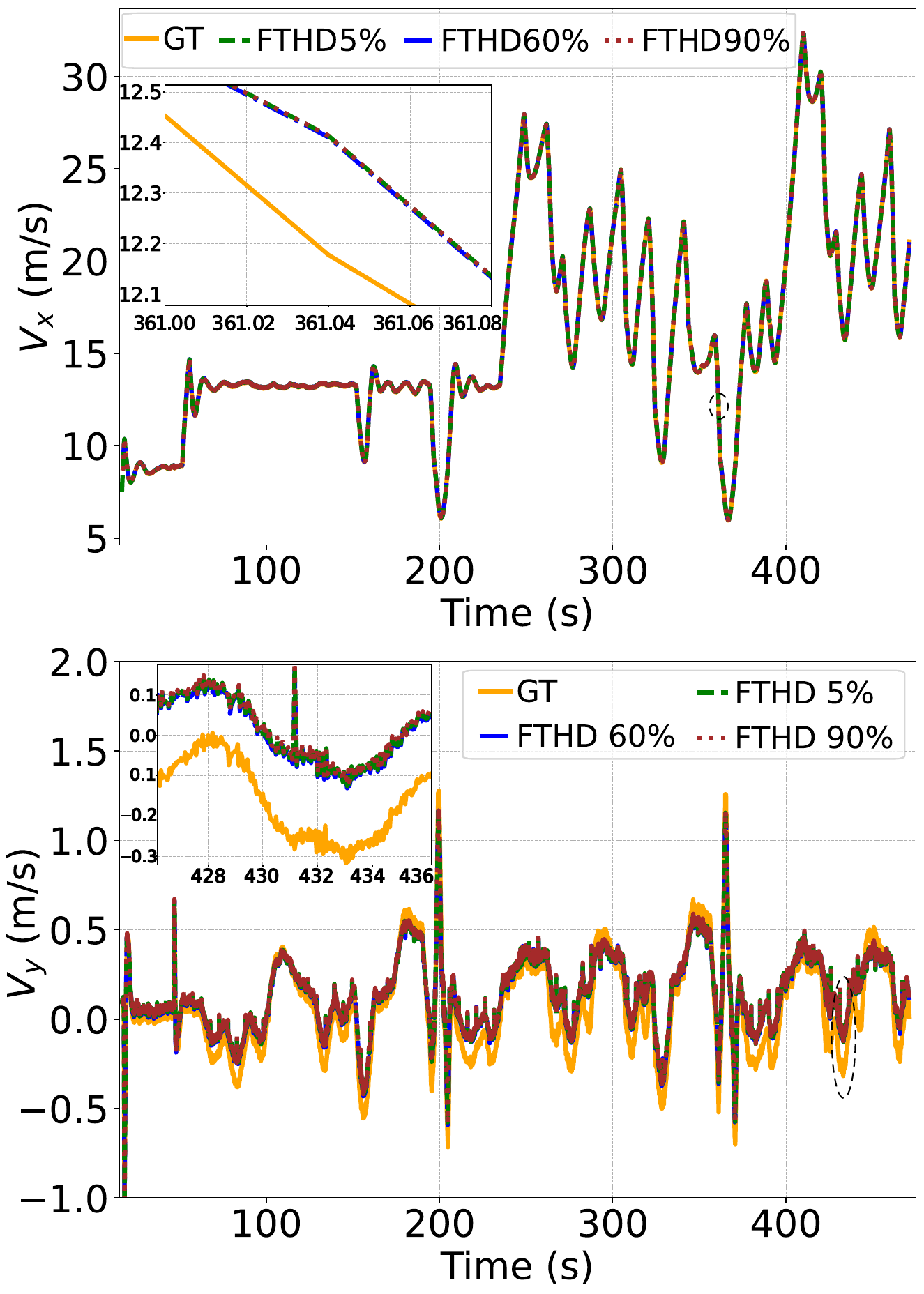}
	\caption{Velocity prediction of FTHD in real experiments as the training set size decreases. (inset) he zoomed-in view highlights the maximum error compared to the GT at the time marked using black dashed circles in the figures. FTHD model demonstrates better overall accuracy in prediction.}
	\label{fig:Real_velocity_FTHD}
\end{figure}

Since the GT for vehicle parameters is unknown in real-world experiments, it becomes difficult to assess the model's quality solely based on force plots. Additionally, due to variations in the dynamic coefficients of the vehicle over time, such as tire wear and changing road conditions, it is challenging to describe all situations using a single set of parameters. To better illustrate this issue, rather than relying on a constant tire model, we present the dynamic force and velocity responses for both the FTHD and DDM models. Fig.~\ref{fig:Real_force_DDM} and Fig.~\ref{fig:Real_force_FTHD} depict the force response as the training dataset size decreases from \(90\%\) to \(5\%\). Fig.~\ref{fig:Real_force_DDM} shows the DDM model trained and tested using data from \(\mathbb{D}^\text{exp}_\text{total}\), while Fig.~\ref{fig:Real_force_FTHD} displays the FTHD model trained and tested with data from \(\mathbb{D}^\text{EKF}\). As the training set size decreases, the DDM's force plot deviates significantly at larger slip angles, whereas the FTHD model maintains a more consistent result, even with only 5\% of the training data - closely resembling the plot from 90\%. Fig.~\ref{fig:Real_velocity_DDM} and Fig.~\ref{fig:Real_velocity_FTHD} show the velocity predictions of DDM and FTHD for real-world data. The FTHD predictions exhibit smaller errors compared to the GT velocities. Table~\ref{tab:RMSEValLossMaxE in Real} compares the minimum validation loss ($L_\text{min}$), the RMSE for velocities, and the maximum velocity errors (\(\epsilon_\text{max}\)) for \(v_x\), \(v_y\), and \(\omega\), for different training data ratios from \(\mathbb{D}^\text{exp}_\text{total}\) and \(\mathbb{D}^\text{EKF}\). Even with the same number of training iterations (2k in total) for both models, the FTHD model consistently outperforms the supervised-only PINN DDM in predicting \(v_x\) and \(v_y\), with the prediction of \(\omega\) showing only minor differences. Notably, even as the training set size decreases to \(5\%\), the FTHD model maintains more accuracy than DDM. %, despite an small increase in the error for $\omega$.

\renewcommand{\arraystretch}{1.2}
\begin{table}[!t]
\caption{Comparison of RMSE, validation loss, and maximum velocity errors for FTHD and DDM in simulation experiments across different training set sizes}
\label{tab:RMSEValLossMaxE}
	\vspace{-3mm}
\centering
\setlength\tabcolsep{5pt}
\begin{tabular*}{\columnwidth}{c|c|cccc}
\hline
\hline
\multicolumn{1}{c|}{\multirow{2}{*}{\(\frac{\mathbb{D}^\text{sim}_\text{train}}{\mathbb{D}^\text{sim}_\text{total}}\)}} & & \multicolumn{2}{c|}{DDM} & \multicolumn{2}{c}{FTHD} \\ \cline{3-6} 
\multicolumn{1}{c|}{}  &                & \multicolumn{1}{c}{RMSE}      &\multicolumn{1}{c|}{\(\epsilon_\text{max}\)}     & \multicolumn{1}{c}{RMSE}      & \multicolumn{1}{c}{\(\epsilon_\text{max}\)}      \\ \hline
\multirow{4}{*}{\(\textbf{80\%}\)} & \(v_x\)        & 6.70$\times 10^{-6}$  &\multicolumn{1}{c|}{2.83$\times 10^{-5}$}   & 5.96$\times 10^{-6}$    & 1.53$\times 10^{-5}$     \\  
                          & \(v_y\)        & 1.96$\times 10^{-5}$  &\multicolumn{1}{c|}{3.61$\times 10^{-5}$}     & 3.07$\times 10^{-6}$    & 1.22$\times 10^{-5}$    \\  
                          & \(\omega\)     & 8.19$\times 10^{-5}$    &\multicolumn{1}{c|}{5.07$\times 10^{-4}$}     & 3.82$\times 10^{-5}$    & 2.30$\times 10^{-4}$    \\ 
                          & \multicolumn{1}{c|}{\(L_\text{min}\)} & \multicolumn{2}{c|}{2.38$\times 10^{-9}$} & \multicolumn{2}{c}{5.01$\times 10^{-10}$}  \\ \hline
\multirow{4}{*}{\(\textbf{50\%}\)} & \(v_x\)        & 2.49$\times 10^{-5}$  &\multicolumn{1}{c|}{6.63$\times 10^{-5}$}   & 1.70$\times 10^{-5}$    & 1.18$\times 10^{-5}$    \\  
                          & \(v_y\)        & 1.51$\times 10^{-5}$  &\multicolumn{1}{c|}{5.34$\times 10^{-5}$}     & 1.68$\times 10^{-5}$    & 3.35$\times 10^{-5}$    \\  
                          & \(\omega\)     & 1.44$\times 10^{-4}$    &\multicolumn{1}{c|}{1.32$\times 10^{-3}$}     & 4.83$\times 10^{-5}$    & 2.89$\times 10^{-4}$    \\ 
                          & \multicolumn{1}{c|}{\(L_\text{min}\)} & \multicolumn{2}{c|}{7.15$\times 10^{-9}$} & \multicolumn{2}{c}{9.68$\times 10^{-10}$}  \\ \hline
\multirow{4}{*}{\(\textbf{30\%}\)} & \(v_x\)        & 3.76$\times 10^{-5}$  &\multicolumn{1}{c|}{1.50$\times 10^{-4}$}   & 7.46$\times 10^{-6}$    & 2.07$\times 10^{-5}$    \\  
                          & \(v_y\)        & 8.61$\times 10^{-5}$  &\multicolumn{1}{c|}{4.84$\times 10^{-4}$}     & 2.30$\times 10^{-5}$    & 7.33$\times 10^{-5}$    \\  
                          & \(\omega\)     & 8.66$\times 10^{-4}$    &\multicolumn{1}{c|}{1.79$\times 10^{-2}$}     & 1.02$\times 10^{-4}$    & 6.16$\times 10^{-4}$    \\ 
                          & \multicolumn{1}{c|}{\(L_\text{min}\)} & \multicolumn{2}{c|}{2.53$\times 10^{-7}$} & \multicolumn{2}{c}{3.67$\times 10^{-9}$}  \\ \hline
\multirow{4}{*}{\(\textbf{20\%}\)} & \(v_x\)        & 1.16$\times 10^{-5}$  &\multicolumn{1}{c|}{8.60$\times 10^{-5}$}   & 1.14$\times 10^{-5}$    & 5.03$\times 10^{-5}$    \\  
                          & \(v_y\)        & 8.46$\times 10^{-5}$  &\multicolumn{1}{c|}{3.69$\times 10^{-4}$}     & 1.50$\times 10^{-5}$    & 5.33$\times 10^{-5}$    \\  
                          & \(\omega\)     & 6.48$\times 10^{-4}$    &\multicolumn{1}{c|}{6.52$\times 10^{-3}$}     & 1.29$\times 10^{-4}$    & 1.22$\times 10^{-3}$    \\ 
                          & \multicolumn{1}{c|}{\(L_\text{min}\)} & \multicolumn{2}{c|}{1.42$\times 10^{-7}$} & \multicolumn{2}{c}{5.68$\times 10^{-9}$}  \\ \hline
\multirow{4}{*}{\(\textbf{15\%}\)} & \(v_x\)        & 8.60$\times 10^{-5}$  &\multicolumn{1}{c|}{3.50$\times 10^{-4}$}   & 4.25$\times 10^{-5}$    & 2.35$\times 10^{-4}$    \\  
                          & \(v_y\)        & 4.99$\times 10^{-4}$  &\multicolumn{1}{c|}{1.48$\times 10^{-3}$}     & 1.38$\times 10^{-4}$    & 6.68$\times 10^{-4}$    \\  
                          & \(\omega\)     & 1.54$\times 10^{-3}$    &\multicolumn{1}{c|}{1.35$\times 10^{-2}$}     & 4.22$\times 10^{-4}$    & 2.92$\times 10^{-3}$    \\ 
                          & \multicolumn{1}{c|}{\(L_\text{min}\)} & \multicolumn{2}{c|}{8.74$\times 10^{-7}$} & \multicolumn{2}{c}{6.64$\times 10^{-8}$}  \\ \hline\hline
\end{tabular*}
\end{table}

\renewcommand{\arraystretch}{1.2}
\begin{table}[!t]
\caption{Comparison of RMSE, validation loss, and maximum velocity errors for FTHD and DDM in real-world experiments across different training set sizes}
\label{tab:RMSEValLossMaxE in Real}
	\vspace{-3mm}
\centering
\setlength\tabcolsep{2.8pt}
\begin{tabular*}{\columnwidth}{c|c|cccc}
\hline \hline
\multicolumn{1}{c|}{{Training}} & & \multicolumn{2}{c|}{DDM (\(\mathbb{D}^\text{exp}_\text{total}\))} & \multicolumn{2}{c}{FTHD (\(\mathbb{D}^\text{EKF}\))} \\ \cline{3-6} 
\multicolumn{1}{c|}{Set Ratio}  &                & \multicolumn{1}{c}{RMSE}      &\multicolumn{1}{c|}{\(\epsilon_\text{max}\)}     & \multicolumn{1}{c}{RMSE}      & \multicolumn{1}{c}{\(\epsilon_\text{max}\)}      \\ \hline
\multirow{4}{*}{\(\textbf{90\%}\)} & \(v_x\)        & 1.852$\times 10^{-2}$  &\multicolumn{1}{c|}{2.361$\times 10^{-1}$}   & 1.315$\times 10^{-2}$    & 1.583$\times 10^{-1}$     \\  
                          & \(v_y\)        & 8.471$\times 10^{-3}$  &\multicolumn{1}{c|}{1.737$\times 10^{-1}$}     & 4.582$\times 10^{-3}$    & 7.309$\times 10^{-2}$    \\  
                          & \(\omega\)     & 3.275$\times 10^{-3}$    &\multicolumn{1}{c|}{6.603$\times 10^{-2}$}     & 3.098$\times 10^{-3}$    & 5.585$\times 10^{-2}$    \\ 
                          & \multicolumn{1}{c|}{\(L_\text{min}\)} & \multicolumn{2}{c|}{1.48$\times 10^{-4}$} & \multicolumn{2}{c}{6.974$\times 10^{-5}$}  \\ \hline
\multirow{4}{*}{\(\textbf{60\%}\)} & \(v_x\)        & 2.181$\times 10^{-2}$  &\multicolumn{1}{c|}{2.361$\times 10^{-1}$}   & 1.64$\times 10^{-2}$    & 1.583$\times 10^{-1}$    \\  
                          & \(v_y\)        & 1.227$\times 10^{-2}$  &\multicolumn{1}{c|}{1.737$\times 10^{-1}$}     & 8.874$\times 10^{-3}$    & 9.951$\times 10^{-2}$    \\  
                          & \(\omega\)     & 3.641$\times 10^{-3}$    &\multicolumn{1}{c|}{6.37$\times 10^{-2}$}     & 3.373$\times 10^{-3}$   & 6.295$\times 10^{-2}$    \\ 
                          & \multicolumn{1}{c|}{\(L_\text{min}\)} & \multicolumn{2}{c|}{2.132$\times 10^{-4}$} & \multicolumn{2}{c}{1.251$\times 10^{-4}$}  \\ \hline
\multirow{4}{*}{\(\textbf{30\%}\)} & \(v_x\)        & 2.526$\times 10^{-2}$  &\multicolumn{1}{c|}{2.736$\times 10^{-1}$}   & 1.95$\times 10^{-2}$    & 1.658$\times 10^{-1}$    \\  
                          & \(v_y\)        & 1.469$\times 10^{-2}$  &\multicolumn{1}{c|}{1.754$\times 10^{-1}$}     & 1.262$\times 10^{-2}$    & 1.269$\times 10^{-1}$    \\  
                          & \(\omega\)     & 4.753$\times 10^{-3}$    &\multicolumn{1}{c|}{6.808$\times 10^{-2}$}     & 4.679$\times 10^{-3}$    & 6.707$\times 10^{-2}$    \\ 
                          & \multicolumn{1}{c|}{\(L_\text{min}\)} & \multicolumn{2}{c|}{2.922$\times 10^{-4}$} & \multicolumn{2}{c}{1.889$\times 10^{-4}$}  \\ \hline
\multirow{4}{*}{\(\textbf{15\%}\)} & \(v_x\)        & 2.628$\times 10^{-2}$  &\multicolumn{1}{c|}{2.577$\times 10^{-1}$}   & 2.055$\times 10^{-2}$    & 1.733$\times 10^{-1}$    \\  
                          & \(v_y\)        & 1.644$\times 10^{-2}$  &\multicolumn{1}{c|}{1.777$\times 10^{-1}$}     & 1.378$\times 10^{-2}$    & 1.533$\times 10^{-1}$    \\  
                          & \(\omega\)     & 5.109$\times 10^{-3}$    &\multicolumn{1}{c|}{7.429$\times 10^{-2}$}     & 4.553$\times 10^{-3}$    & 7.483$\times 10^{-2}$    \\ 
                          & \multicolumn{1}{c|}{\(L_\text{min}\)} & \multicolumn{2}{c|}{3.29$\times 10^{-4}$} & \multicolumn{2}{c}{2.135$\times 10^{-4}$}  \\ \hline
\multirow{4}{*}{\(\textbf{5\%}\)} & \(v_x\)        & 2.795$\times 10^{-2}$  &\multicolumn{1}{c|}{3.101$\times 10^{-1}$}   & 2.204$\times 10^{-2}$    & 2.135$\times 10^{-1}$    \\  
                          & \(v_y\)        & 1.846$\times 10^{-2}$  &\multicolumn{1}{c|}{1.799$\times 10^{-1}$}     & 1.765$\times 10^{-2}$    & 1.285$\times 10^{-1}$    \\  
                          & \(\omega\)     & 5.958$\times 10^{-3}$    &\multicolumn{1}{c|}{1.293$\times 10^{-1}$}     & 4.806$\times 10^{-3}$    & 7.628$\times 10^{-2}$    \\ 
                          & \multicolumn{1}{c|}{\(L_\text{min}\)} & \multicolumn{2}{c|}{3.859$\times 10^{-4}$} & \multicolumn{2}{c}{2.773$\times 10^{-4}$}  \\ \hline \hline
\end{tabular*}
\end{table}

\subsection{Hyperparameters Tuning}
\label{Hyperparameters}
To ensure a fair comparison without the influence of model hyperparameters, we utilize Tune~\cite{liaw2018tune} and set up identical configuration spaces and trials for both models. This includes the selection of hidden layers, GRU layers, neurons, learning rate, the choice of \(N\) for each input features \(\textbf{X}_\text{input}\), and batch size, as well as the frozen layers during fine-tuning process. To simplify the process of hyperparameters selection, we set the size of frozen layers equal to the $3/4$ of the total layers for each tests. Both models are trained on the same hardware: a GeForce RTX 4090 GPU, a 13th Gen Intel Core i7-13700K CPU, and 128 GB of RAM. To further challenge the model's performance, the full training set is used for validation. The lowest validation loss and the corresponding \(\Phi\) are used for the simulation comparisons. In the real experiment, we use the EKF-FTHD model with the lowest validation loss to generate the filtered dataset used for training the FTHD estimation model, then select the hyperparameter configuration with the lowest validation loss for dynamical performance comparisons. Tables~\ref{tab:sim_hyperparameters} and~\ref{tab:real_hyperparameters} show the model configurations where the lowest validation loss occurs after reducing the training dataset size. In the simulations, each configuration represents both DDM and FTHD with the same original dataset. In the real experiments, each configuration represents the DDM model with the original dataset and the FTHD model with the filtered dataset.

% Fig.~\ref{fig:HyperTuning} illustrates the hyperparameter tuning trials for the FTHD. The ``Validation Loss" layer represents the lowest validation loss for each combination, and the yellow line highlights the combination that points to the smallest one among all trials.

% \begin{figure}[ht]
%   \centering
%   \includegraphics[width=\linewidth]{photo/HyperparameterTuning2.pdf}
%   \caption{A visualization of the hyperparameter tuning trials, with the best FTHD highlighted in yellow.}
%   \label{fig:HyperTuning}
% \end{figure}

\renewcommand{\arraystretch}{1.5}
\begin{table}[!t]
\caption{Hyperparameters used in simulation experiments for DDM and FTHD models, showing configurations where the lowest validation loss occurs as the training dataset size decreases}
\centering
	\vspace{-3mm}
\label{tab:sim_hyperparameters}
\setlength\tabcolsep{5pt}
\begin{tabular*}{\columnwidth}{c|c|c|c|c|c|c}
\hline
\hline
\multicolumn{1}{c|}{\multirow{2}{*}{\(\frac{\mathbb{D}^\text{train}}{\mathbb{D}^\text{sim}_\text{total}}\)}} & \multicolumn{1}{c|}{\multirow{2}{*}{Hidden}} & \multicolumn{1}{c|}{\multirow{2}{*}{Learning}} & \multicolumn{1}{c|}{\multirow{2}{*}{GRU}} & \multicolumn{1}{c|}{\multirow{2}{*}{Hidden}} & \multicolumn{1}{c|}{\multirow{2}{*}{History}} & \multicolumn{1}{c}{\multirow{2}{*}{Batch}} \\
\multicolumn{1}{c|}{} & \multicolumn{1}{c|}{Layers}  &\multicolumn{1}{c|}{Rate} &\multicolumn{1}{c|}{Layers}&\multicolumn{1}{c|}{Layer size} &\multicolumn{1}{c|}{timesteps} &\multicolumn{1}{c}{size}\\ \hline
\multicolumn{1}{c|}{$\textbf{80\%}$} & \multicolumn{1}{c|}{8} & \multicolumn{1}{c|}{0.002510} & \multicolumn{1}{c|}{3} & \multicolumn{1}{c|}{49} & \multicolumn{1}{c|}{9} & \multicolumn{1}{c}{32} \\ \hline
\multicolumn{1}{c|}{$\textbf{50\%}$} & \multicolumn{1}{c|}{6} & \multicolumn{1}{c|}{0.000734} & \multicolumn{1}{c|}{4} & \multicolumn{1}{c|}{230} & \multicolumn{1}{c|}{7} & \multicolumn{1}{c}{32} \\ \hline
\multicolumn{1}{c|}{$\textbf{30\%}$} & \multicolumn{1}{c|}{7} & \multicolumn{1}{c|}{0.000898} & \multicolumn{1}{c|}{3} & \multicolumn{1}{c|}{164} & \multicolumn{1}{c|}{3} & \multicolumn{1}{c}{32} \\ \hline
\multicolumn{1}{c|}{$\textbf{20\%}$} & \multicolumn{1}{c|}{8} & \multicolumn{1}{c|}{0.000985} & \multicolumn{1}{c|}{0} & \multicolumn{1}{c|}{88} & \multicolumn{1}{c|}{18} & \multicolumn{1}{c}{32} \\ \hline
\multicolumn{1}{c|}{$\textbf{15\%}$} & \multicolumn{1}{c|}{5} & \multicolumn{1}{c|}{0.003907} & \multicolumn{1}{c|}{1} & \multicolumn{1}{c|}{25} & \multicolumn{1}{c|}{18} & \multicolumn{1}{c}{32} \\ \hline\hline
\end{tabular*}
\end{table}

\renewcommand{\arraystretch}{1.5}
\begin{table}[!t]
\caption{Hyperparameters used in real-world experiments for DDM and FTHD models, indicating configurations where the lowest validation loss occurs as the training dataset size decreases}
\centering
	\vspace{-3mm}
\label{tab:real_hyperparameters}
\setlength\tabcolsep{4.2pt}
\begin{tabular*}{\columnwidth}{c|c|c|c|c|c|c}
\hline
\hline
\multicolumn{1}{c|}{\multirow{2}{*}{Training}} & \multicolumn{1}{c|}{\multirow{2}{*}{Hidden}} & \multicolumn{1}{c|}{\multirow{2}{*}{Learning}} & \multicolumn{1}{c|}{\multirow{2}{*}{GRU}} & \multicolumn{1}{c|}{\multirow{2}{*}{Hidden}} & \multicolumn{1}{c|}{\multirow{2}{*}{History}} & \multicolumn{1}{c}{\multirow{2}{*}{Batch}} \\
\multicolumn{1}{c|}{Set Ratio} & \multicolumn{1}{c|}{Layers}  &\multicolumn{1}{c|}{Rate} &\multicolumn{1}{c|}{Layers}&\multicolumn{1}{c|}{Layer size} &\multicolumn{1}{c|}{timesteps} &\multicolumn{1}{c}{size}\\ \hline
\multicolumn{1}{c|}{$\textbf{90\%}$} & \multicolumn{1}{c|}{4} & \multicolumn{1}{c|}{0.001378} & \multicolumn{1}{c|}{2} & \multicolumn{1}{c|}{146} & \multicolumn{1}{c|}{16} & \multicolumn{1}{c}{64} \\ \hline
\multicolumn{1}{c|}{$\textbf{60\%}$} & \multicolumn{1}{c|}{2} & \multicolumn{1}{c|}{0.008942} & \multicolumn{1}{c|}{3} & \multicolumn{1}{c|}{147} & \multicolumn{1}{c|}{18} & \multicolumn{1}{c}{128} \\ \hline
\multicolumn{1}{c|}{$\textbf{30\%}$} & \multicolumn{1}{c|}{4} & \multicolumn{1}{c|}{0.001904} & \multicolumn{1}{c|}{1} & \multicolumn{1}{c|}{233} & \multicolumn{1}{c|}{10} & \multicolumn{1}{c}{64} \\ \hline
\multicolumn{1}{c|}{$\textbf{15\%}$} & \multicolumn{1}{c|}{3} & \multicolumn{1}{c|}{0.002941} & \multicolumn{1}{c|}{2} & \multicolumn{1}{c|}{208} & \multicolumn{1}{c|}{14} & \multicolumn{1}{c}{64} \\ \hline
\multicolumn{1}{c|}{$\textbf{5\%}$} & \multicolumn{1}{c|}{6} & \multicolumn{1}{c|}{0.004099} & \multicolumn{1}{c|}{0} & \multicolumn{1}{c|}{128} & \multicolumn{1}{c|}{4} & \multicolumn{1}{c}{128} \\ \hline \hline
\end{tabular*}
\end{table}
\section{Conclusion}
\label{sec:concl}

In this paper, we introduce FTHD, a novel fine-tuning PINN model that integrates hybrid loss functions to enhance vehicle dynamics estimation. FTHD builds upon pre-trained DDM models, providing an innovative framework that leverages both data-driven and differential loss functions to refine vehicle dynamic coefficient estimation. By addressing the limitations of the sigmoid guard layer with fine-tuning and unsupervised differential loss, FTHD effectively overcomes issues such as local minima, especially with smaller training datasets. The combination of supervised data-driven loss and unsupervised differential loss allows FTHD to consistently outperform DDM in force estimation, particularly as training data size decreases, leading to reduced validation losses and smaller absolute differences between estimated dynamics parameters and ground truth.
Additionally, we propose EKF-FTHD, a model that integrates an EKF to preprocess noisy real-world data while preserving essential physical relationships between states and controls. EKF-FTHD addresses the challenge of unknown ground truth vehicle parameters and ranges, allowing the filtered data to improve model performance even in DDM. The results demonstrate that FTHD offers significant improvements in accuracy and stability when estimating vehicle dynamics, particularly in scenarios with limited training data, making it a robust solution for real-world applications.

Future research will focus on two key areas. First, expanding the EKF-FTHD approach from the bicycle model to more complex systems, such as four-wheel vehicle models, could improve accuracy and applicability by better addressing real-world measurement errors. This method could also be adapted for other dynamic systems beyond vehicle dynamics, including submarines or agricultural machinery, where estimating internal parameters is challenging. Second, integrating FTHD into the design of distributionally robust control systems could enhance controller performance under uncertainty. By using the noise separated by EKF-FTHD as part of an ambiguity model and eventually incorporating it into the constraints, this approach could advance the development of distributionally robust optimization (DRO) controllers that effectively manage uncertainty and deliver robust performance under varying conditions.

% \section*{Acknowledgment}

% \ifCLASSOPTIONcaptionsoff
%   \newpage
% \fi



%\begin{thebibliography}{1}


\bibliographystyle{IEEEtran}
\bibliography{KaiyanYuRef,Bibliography}

\begin{thebibliography}{10}
\providecommand{\url}[1]{#1}
\csname url@rmstyle\endcsname
\providecommand{\newblock}{\relax}
\providecommand{\bibinfo}[2]{#2}
\providecommand\BIBentrySTDinterwordspacing{\spaceskip=0pt\relax}
\providecommand\BIBentryALTinterwordstretchfactor{4}
\providecommand\BIBentryALTinterwordspacing{\spaceskip=\fontdimen2\font plus
\BIBentryALTinterwordstretchfactor\fontdimen3\font minus
  \fontdimen4\font\relax}
\providecommand\BIBforeignlanguage[2]{{%
\expandafter\ifx\csname l@#1\endcsname\relax
\typeout{** WARNING: IEEEtran.bst: No hyphenation pattern has been}%
\typeout{** loaded for the language `#1'. Using the pattern for}%
\typeout{** the default language instead.}%
\else
\language=\csname l@#1\endcsname
\fi
#2}}

\bibitem{bakker1989new}
E.~Bakker, H.~B. Pacejka, and L.~Lidner, ``A new tire model with an application
  in vehicle dynamics studies,'' \emph{SAE Trans.}, vol.~98, no.~6, pp.
  101--113, 1989.

\bibitem{weiss2020deepracing}
T.~Weiss and M.~Behl, ``{DeepRacing}: A framework for autonomous racing,''
  Grenoble, France, 2020, pp. 1163--1168.

\bibitem{kim2022physics}
T.~Kim, H.~Lee, and W.~Lee, ``Physics embedded neural network vehicle model and
  applications in risk-aware autonomous driving using latent features,'' in
  \emph{Proc. IEEE/RSJ Int. Conf. Intell. Robot. Syst.}, Kyoto, Japan, 2022,
  pp. 4182--4189.

\bibitem{chrosniak2023deep}
J.~Chrosniak, J.~Ning, and M.~Behl, ``Deep dynamics: Vehicle dynamics modeling
  with a physics-constrained neural network for autonomous racing,''
  \emph{{IEEE} Robot. Autom. Lett.}, vol.~9, no.~5, pp. 5292--5297, 2024.

\bibitem{FangMECC2024}
S.~Fang and K.~Yu, ``{Fine-tuning hybrid physics-informed neural networks for
  vehicle dynamics model estimation},'' in \emph{Proc. Modeling Estimation
  Control Conf.}, Chicago, IL, USA, 2024, to appear.

\bibitem{vicente2020linear}
B.~A.~H. Vicente, S.~S. James, and S.~R. Anderson, ``Linear system
  identification versus physical modeling of lateral--longitudinal vehicle
  dynamics,'' \emph{{IEEE} Trans. Contr. Syst. Technol.}, vol.~29, no.~3, pp.
  1380--1387, 2020.

\bibitem{dias2014longitudinal}
J.~E.~A. Dias, G.~A.~S. Pereira, and R.~M. Palhares, ``Longitudinal model
  identification and velocity control of an autonomous car,'' \emph{{IEEE}
  Trans. Intell. Transport. Syst.}, vol.~16, no.~2, pp. 776--786, 2014.

\bibitem{vyasarayani2011parameter}
C.~P. Vyasarayani, T.~Uchida, A.~Carvalho, and J.~McPhee, ``Parameter
  identification in dynamic systems using the homotopy optimization approach,''
  \emph{Multibody Syst. Dyn.}, vol.~26, pp. 411--424, 2011.

\bibitem{JainRaceOpt2020}
A.~Jain and M.~Morari, ``Computing the racing line using {Bayesian}
  optimization,'' in \emph{Proc. IEEE Conf. Decision Control}, Jeju Island,
  Republic of Korea, 2020, pp. 6192--6197.

\bibitem{kabzan2019learning}
J.~Kabzan, L.~Hewing, A.~Liniger, and M.~N. Zeilinger, ``Learning-based model
  predictive control for autonomous racing,'' \emph{{IEEE} Robot. Autom.
  Lett.}, vol.~4, no.~4, pp. 3363--3370, 2019.

\bibitem{spielberg2019neural}
N.~A. Spielberg, M.~Brown, N.~R. Kapania, J.~C. Kegelman, and J.~C. Gerdes,
  ``Neural network vehicle models for high-performance automated driving,''
  vol.~4, no.~28, p. eaaw1975, 2019.

\bibitem{williams2017information}
G.~Williams, N.~Wagener, B.~Goldfain, P.~Drews, J.~M. Rehg, B.~Boots, and E.~A.
  Theodorou, ``Information theoretic mpc for model-based reinforcement
  learning,'' in \emph{Proc. IEEE Int. Conf. Robot. Autom.}, Singapore, 2017,
  pp. 1714--1721.

\bibitem{hermansdorfer2021end}
L.~Hermansdorfer, R.~Trauth, J.~Betz, and M.~Lienkamp, ``End-to-end neural
  network for vehicle dynamics modeling,'' Agadir, Morocco, 2021, pp. 407--412.

\bibitem{xu2022physics}
P.-F. Xu, C.-B. Han, H.-X. Cheng, C.~Cheng, and T.~Ge, ``A physics-informed
  neural network for the prediction of unmanned surface vehicle dynamics,''
  vol.~10, no.~2, p. 148, 2022.

\bibitem{koysuren2023online}
K.~Koysuren, A.~F. Keles, and M.~Cakmakci, ``Online parameter estimation using
  physics-informed deep learning for vehicle stability algorithms,'' in
  \emph{Proc. Amer. Control Conf.}, Denver, CO, USA, 2023, pp. 466--471.

\bibitem{schafer2011savitzky}
R.~W. Schafer, ``What is a savitzky-golay filter?[lecture notes],'' \emph{IEEE
  Signal Process. Mag.}, vol.~28, no.~4, pp. 111--117, 2011.

\bibitem{ribeiro2004kalman}
M.~I. Ribeiro, ``Kalman and extended kalman filters: Concept, derivation and
  properties,'' \emph{Institute for Systems and Robotics}, vol.~43, no.~46, pp.
  3736--3741, 2004.

\bibitem{wischnewski2022indy}
A.~Wischnewski, M.~Geisslinger, J.~Betz, \emph{et~al.}, ``Indy autonomous
  challenge - autonomous race cars at the handling limits,'' in \emph{Proc.
  Int. Munich Chassis Symp.}, Munich, Germany, 2022, pp. 163--182.

\bibitem{o2019f1}
M.~O'Kelly, V.~Sukhil, H.~Abbas, \emph{et~al.}, ``F1/10: An open-source
  autonomous cyber-physical platform,'' \emph{arXiv:1901.08567, {\tt
  www.arxiv.org}}, 2019.

\bibitem{pacejka1992magic}
H.~B. Pacejka and E.~Bakker, ``The magic formula tyre model,'' \emph{Vehicle
  system dynamics}, vol.~21, no.~S1, pp. 1--18, 1992.

\bibitem{liu2021dual}
D.~Liu and Y.~Wang, ``A dual-dimer method for training physics-constrained
  neural networks with minimax architecture,'' \emph{Neural Netw.}, vol. 136,
  pp. 112--125, 2021.

\bibitem{lew2022physics}
C.~Lew, ``Physics informed neural networks: Reducing data size requirements via
  hybrid learning,'' Vancouver, WA, USA, 2022, pp. 178--179.

\bibitem{swirszcz2016local}
G.~Swirszcz, W.~M. Czarnecki, and R.~Pascanu, ``Local minima in training of
  neural networks,'' \emph{arXiv:1611.06310, {\tt www.arxiv.org}}, 2016.

\bibitem{kulkarni2023racecar}
A.~Kulkarni, J.~Chrosniak, E.~Ducote, F.~Sauerbeck, A.~Saba, U.~Chirimar,
  J.~Link, M.~Behl, and M.~Cellina, ``Racecar-the dataset for high-speed
  autonomous racing,'' in \emph{Proc. IEEE/RSJ Int. Conf. Intell. Robot.
  Syst.}, Detroit, MI, USA., 2023, pp. 11\,458--11\,463.

\bibitem{liaw2018tune}
R.~Liaw, E.~Liang, R.~Nishihara, P.~Moritz, J.~E. Gonzalez, and I.~Stoica,
  ``Tune: A research platform for distributed model selection and training,''
  \emph{arXiv:1807.05118, {\tt www.arxiv.org}}, 2018.

\end{thebibliography}
%\vskip 0pt plus -1fil
\begin{biography}[{\includegraphics[width=1in,height=1.25in,clip,keepaspectratio]{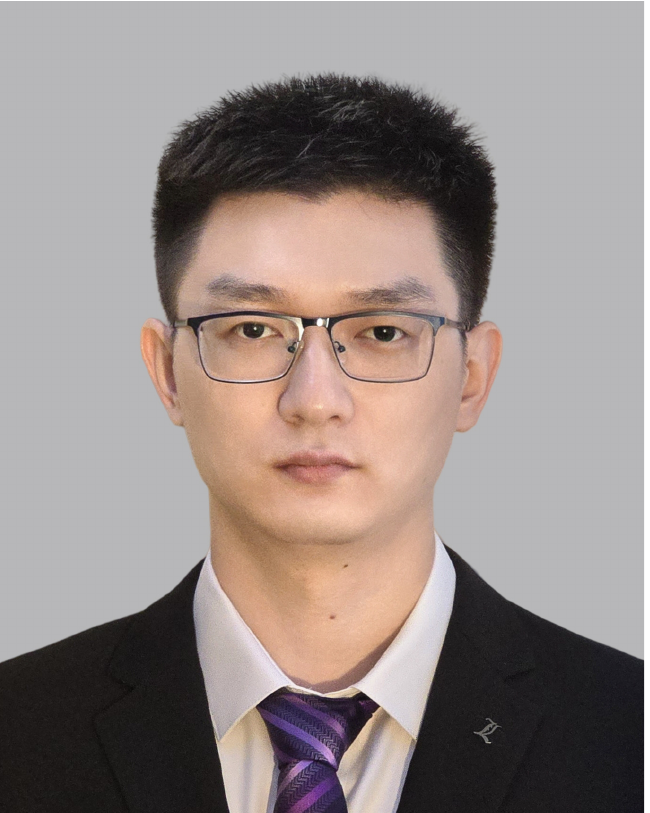}}]{Shiming Fang} (Student Member, IEEE) received the B.S. degree in Mechanical Engineering from Wuhan University of Technology, Wuhan, China in 2016, and the M.S. degree in Mechanical Engineering from University of Birmingham, Birmingham, UK in 2017. He is currently working towards a Ph.D. degree in Mechanical Engineering at Binghamton University. His current research focuses on autonomous vehicle modeling, robust control, and artificial intelligence in autonomous driving. 	
\end{biography}
%\vskip 0pt plus -1fil

\vspace{-3\baselineskip}
\begin{biography}[{\includegraphics[width=1in,height=1.25in,clip,keepaspectratio]{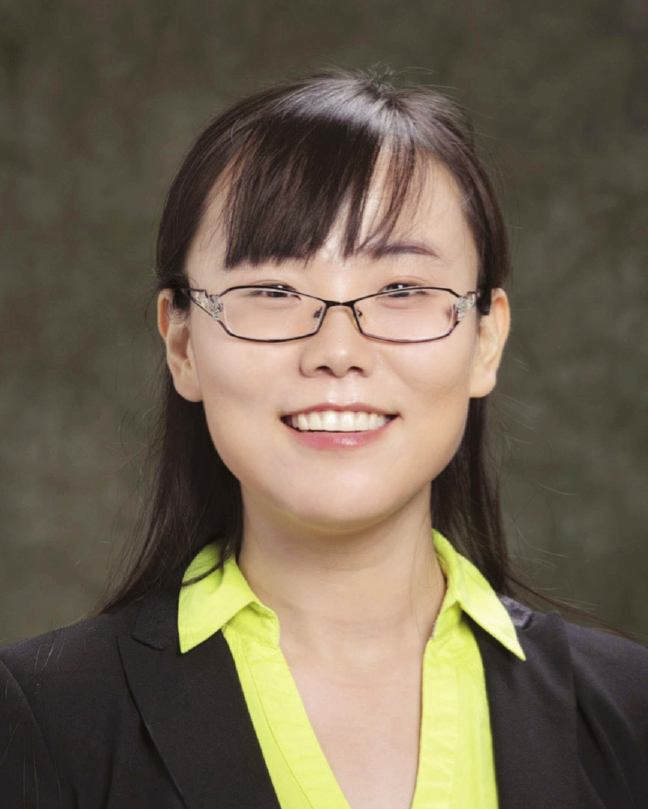}}]{Kaiyan Yu} (Member, IEEE) received the B.S. degree in intelligent science and technology from Nankai University, Tianjin, China, in 2010, and the Ph.D. degree in mechanical and aerospace engineering from Rutgers University, Piscataway, NJ, USA, in 2017. She joined the Department of Mechanical Engineering, Binghamton University, Binghamton, NY, USA, in 2018, where she is currently an Assistant Professor. Her current research interests include autonomous robotic systems, motion planning and control, mechatronics, automation science and engineering with applications to nano/micro particles control and manipulation, Lab-on-a-chip, and biomedical systems. 
	
	Dr. Yu is a member of the American Society of Mechanical Engineers (ASME). She is a recipient of the 2022 US NSF CAREER Award. She currently serves as an Associate Vice President of the {\em IEEE Robotics and Automation Society (RAS) Media Services Board} (since 2019) and an Associate Editor of the {\sc IEEE Transactions on Automation Science and Engineering}, {\em IEEE Robotics and Automation Letters}, {\em IFAC Mechatronics}, {\em Frontiers in Robotics and AI}, and the IEEE Robotics and Automation Society Conference Editorial Board and the ASME Dynamic Systems and Control Division Conference Editorial Board  (since 2018).		
\end{biography}
%\end{thebibliography}
 
\end{document}